\documentclass[11pt]{article}

\usepackage[utf8]{inputenc}
\usepackage[T1]{fontenc}
\usepackage{amsmath,amssymb,amsfonts}
\usepackage{amsthm}
\usepackage{graphicx}
\usepackage{multirow}
\usepackage[title]{appendix}
\usepackage{xcolor}
\usepackage{textcomp}
\usepackage{manyfoot}
\usepackage{booktabs}
\usepackage{algorithm}
\usepackage{algorithmicx}
\usepackage{algpseudocode}
\usepackage{hyperref}
\usepackage{geometry}
\theoremstyle{plain}

\theoremstyle{remark}

\theoremstyle{definition}
\newtheorem{definition}{Definition}

\providecommand{\backmatter}{}

\title{SILVA Networks as Structured Implicit Layers and Vector Attractors via Dynamic Interaction Fields}

\author{
Jose Luis Lima de Jesus Silva*1,2\\[0.4em]
{\small 1 Federal University of Bahia, Department of Geophysics, Salvador, BA 40170-115, Brazil}\\
{\small 2 Grupo de Estudos e Aplica{\c c}{\~a}o de Intelig{\^e}ncia Artificial em Geof{\'i}sica (GAIA),}\\
{\small Federal University of Bahia, Salvador, BA 40170-115, Brazil}\\
{\small * Corresponding author: \href{mailto:jseluis.silva@gmail.com}{jseluis.silva@gmail.com}}
}
\date{}

\begin{document}
\maketitle

\begin{abstract}
Many learning problems require representations that reconcile direct input,
nearby structure, and broader context. In implicit neural layers, these
influences are usually absorbed into a single fixed-point update, making it
hard to identify what enters from the stimulus, what propagates locally, what
comes from global context, and what is produced by solver dynamics. Here we
introduce SILVA Networks, Structured Implicit Layers and Vector Attractors via
Dynamic Interaction Fields. SILVA separates stimulus, local interaction, global
interaction, damping, and readout inside one fixed-point architecture. The same
template is instantiated for images, molecules, citation networks, and
long-range graph benchmarks through domain-specific definitions of nodes,
neighborhoods, and global summaries. Experiments and ablations show
task-dependent roles for these terms: local interactions are load-bearing in
the graph tasks, MNIST gains little from recurrence at the tested capacity, and
the clearest global benefit appears in a long-range node-classification
benchmark. SILVA therefore provides an implicit representation whose internal
interaction dynamics can be trained, ablated, visualized, and diagnosed.
\end{abstract}

% \noindent\textbf{Keywords} deep equilibrium models, graph attention networks, fixed-point iteration, molecular property prediction, node classification, dynamical systems

\section{Introduction}\label{sec1}

Latent representations often have to reconcile several simultaneous sources of
influence. A pixel region, graph node, or molecular atom is driven by its own
observed features, by nearby structure, and by broader context from the system
in which it appears \cite{battaglia2016interaction,gilmer2017neural}. Modern
architectures combine these effects through depth, message passing, attention,
or recurrence \cite{gilmer2017neural,vaswani2017attention}. In an implicit
architecture, the same ingredients can be viewed as a driven dynamical system:
an input stimulus enters a state space, local and global interactions modify
that state, and a numerical solver returns the state reached at convergence
\cite{bai2019deq,gu2020ignn}. The proposed Structured Implicit Layers and
Vector Attractors via Dynamic Interaction Fields (SILVA Networks) architecture
makes this decomposition explicit by separating the stimulus, interaction
operators, solver update, and readout state.

This view continues a long progression from one-pass transformations toward
iterative construction of latent states. Backpropagation made explicitly
layered networks trainable \cite{rumelhart1986backprop}. Hopfield networks and
energy-based models showed that a representation can also be the state reached
by a dynamical system \cite{hopfield1982neural,lecun2006tutorial}. Residual
networks made deep iterative refinement trainable by learning increments around
an identity map \cite{he2016resnet}, while neural ordinary differential
equations recast depth as continuous-time dynamics solved numerically
\cite{chen2018neuralode}. Graph neural networks (GNNs) extended iterative state
updates to relational data by recursively updating node states over edges
\cite{scarselli2009gnn}.

Deep equilibrium models (DEQs) \cite{bai2019deq} make the fixed-point view explicit. Instead of
specifying a finite stack of distinct layers, they define a transformation and
solve for $z^\ast=f_\theta(z^\ast,x)$. In the original DEQ formulation,
implicit differentiation decouples training memory from the number of solver
iterations \cite{bai2019deq}. Graph-specific implicit neural networks extend
this idea to node-state matrices and long-range graph dependencies
\cite{gu2020ignn}, with later operator-theoretic formulations emphasizing
well-posedness, stability, and accelerated fixed-point solvers
\cite{baker2023ignn}. Jacobian-regularized DEQs further connect the conditioning
of the equilibrium Jacobian to forward and backward stability
\cite{bai2021jacobian}. Multiscale DEQs (MDEQs) further show that equilibria can
couple several feature resolutions inside one joint fixed point
\cite{bai2020mdeq}. The fixed-point abstraction establishes a powerful
representation principle. SILVA focuses on the internal composition of that
fixed-point update, specifying how direct input drive, local interaction, and
system-wide context enter the same equilibrium computation.

Several neural architectures already implement interaction mechanisms.
Geometric deep learning treats grids, graphs, and manifolds as one family by
generalizing convolution beyond Euclidean arrays
\cite{bronstein2017geometric}. Graph convolution, message passing, and
interaction networks provide efficient local aggregation over edges or objects
\cite{battaglia2016interaction,kipf2017gcn,gilmer2017neural}. Graph attention
networks (GATs) and transformer attention learn data-dependent interaction weights
\cite{velickovic2018gat,vaswani2017attention}. Purely local propagation,
however, can compress distant signals into fixed-size node states
\cite{alon2021bottleneck}. Dense global attention and latent bottleneck
architectures can also entangle local computation, global context, and repeated
solver application inside a single learned transformation
\cite{jaegle2022perceiverio,bae2022graphperceiverio}. SILVA separates four
computational objects within one fixed-point layer: the external stimulus, the
local interaction operator, the global interaction operator, and the damped
solver that accumulates their effects.

The terminology of stimulus, interaction, stability, and attractor has a long
history in physics-inspired and neurodynamic models. Hopfield networks framed
computation as collective settling in a physical system of interacting units
\cite{hopfield1982neural}. Wilson-Cowan population dynamics and Amari neural
fields model neural activity as a driven interacting population over time
\cite{wilson1972excitatory,amari1977dynamics}. Modern neuroscience uses
attractor networks to explain persistent activity, cue integration, and error
correction in neural populations
\cite{khona2022attractor}. Visual-cortical models further show that tuning and
perception can depend on recurrent, lateral, predictive, and top-down
interactions rather than feedforward drive alone
\cite{benyishai1995theory,rao1999predictive,gilbert2013topdown}. These works
motivate the dynamical vocabulary used here, while SILVA remains a
machine-learning architecture rather than a biological circuit model.

The local--global tradeoff appears differently across benchmarks. The ZINC
molecular-property benchmark often rewards chemically local bond structure
\cite{irwin2012zinc}. CLUSTER, from the Benchmarking GNNs suite, is an
inductive node-classification benchmark over synthetic graph instances in which
long-range aggregation is central \cite{dwivedi2020benchmarking}. Cora,
Citeseer, and Pubmed citation-network node classification sit at another boundary
\cite{sen2008collective,yang2016revisiting}. Their labelled budgets are small
and strong local baselines often suffice, making these datasets a direct test
of whether a global term contributes measurable signal once local structure
already explains much of the task.

\textbf{SILVA Networks} names the mechanism's three components. \emph{Structured
implicit layers} are fixed-point layers
whose update is split into explicit computational roles. \emph{Vector
attractors} are the settled vector-valued states $z^\ast$ reached by the
solver. \emph{Dynamic interaction fields} are learned fields
$f_\theta(z,x)-z$ that combine stimulus, local interaction, and global
interaction. At the layer level, SILVA uses a damped update
\(z_{k+1}=(1-\alpha)z_k+\alpha f_\theta(z_k,x)\), where $f_\theta$ is
organized as stimulus plus local and global terms, with domain-specific
adapters defining the state space and operators.

This formulation generalizes a conventional neural layer in a specific sense.
A standard feedforward transformation computes a representation in one pass.
SILVA instead injects a stimulus and lets it propagate through an interacting
state until an approximate vector attractor is reached. In the locally
linearized analysis developed below, a path is an ordered product of self,
local, and global operators carrying stimulus to a state coordinate. These
stimulus-to-state paths make precise how direct input, local neighborhoods,
global context, damping, and finite solver depth combine. The same outer
fixed-point solver is used across domains, while each domain defines what
counts as a node, a local neighborhood, and a global summary. The shared object
is an inspectable interaction template instantiated through domain-specific
state spaces and operators.

The term equilibrium is used in two precise senses. Architecturally, all models
run a damped fixed-point solver and read out the terminal state $z_K$ as an
approximate fixed point. In the strict DEQ training sense, the adjoint-trained
MNIST diagnostic uses a GMRES adjoint solve \cite{saad1986gmres}, whereas the
remaining experiments use truncated backpropagation through the unrolled
fixed-point solver.

The empirical results are informative because the global term is not uniformly
beneficial. On MNIST \cite{lecun1998gradient} and citation-network node
classification, the global term does not automatically improve accuracy over
local or no-interaction variants, so the added operators are tested directly
rather than assumed to help. On the
long-range CLUSTER benchmark, where broader aggregation is expected to matter,
the full local+global SILVA model reaches
$73.04\pm0.60\%$ validation accuracy across five seeds. This exceeds the best
matched local graph baseline by $14.29$ percentage points and exceeds a
no-global SILVA ablation by $5.49$ percentage points. The same experiments
also expose solver residuals,
spectral-radius behavior, learned interaction structure, and
equilibrium-state geometry, providing measurements of the dynamics as well as
task scores.

\textbf{Contributions.}
\begin{enumerate}
  \item SILVA Networks define structured implicit layers in which a stimulus
  term, a local interaction term, a global interaction term, and a damped
  solver appear as separate parts of one dynamic interaction field
  (Section~\ref{sec:method}).
  \item The same template is instantiated for vision, molecular regression,
  citation-network node classification, and long-range graph benchmarks by
  specifying the state nodes, local graph, global summary, solver, and backward
  mode used in each domain.
  \item Stacked SILVA models compose separately solved approximate vector
  attractor states with distinct parameters and timescales, rather than increasing the iteration
  count of a single equilibrium solve.
  \item A finite path-sum interpretation of the locally linearized solver
  connects solver depth, damping, spectral radius, and effective
  stimulus-to-state influence through the same Jacobian-level operator
  (Sections~\ref{sec:pathsum} and~\ref{sec:green}).
  \item The architecture is evaluated across vision, chemistry,
  citation-network, and long-range graph settings, including local/global
  ablations that identify when the global interaction contributes measurable
  signal and when local structure already dominates.
  \item The empirical characterization includes seed sensitivity, finite solver
  budgets, truncated backpropagation, stability diagnostics, and direct
  benchmark overlap with general-purpose multi-domain architectures.
\end{enumerate}

\section{Method}\label{sec:method}

\subsection{The equilibrium layer}

Let $z \in \mathbb{R}^{N\times d}$ be a layer's state over $N$ entities
(pixels/units for vision, atoms for chemistry, nodes for citation graphs)
with hidden width $d$, and $x$ the layer's input embedding. All experiments
share one outer template. A per-domain update map $f_\theta(z,x)$ reaches a
fixed point $z^\ast=f_\theta(z^\ast,x)$ via damped fixed-point (Picard)
iteration \cite{banach1922operations} for
$k=0,\dots,K-1$,
\begin{equation}
z_{k+1} = (1-\alpha)\, z_k + \alpha \cdot f_\theta(z_k, x),
\label{eq:picard}
\end{equation}
Here $z_k$ is the solver iterate after $k$ updates, $K$ is the total iteration
budget, $\alpha\in(0,1]$ is the damping coefficient, $\theta$ denotes all
learned parameters in the update map, and $z_K$ is the approximate fixed point
used by the downstream readout. The input $x$ is already encoded by the
domain-specific input encoder defined below.
The update map always combines stimulus, local, and global terms. Its exact
composition varies by domain, so the equations below state the evaluated
compositions separately. The local and global maps act on a recurrent signal
$\tilde z=\chi_{\mathcal D}(z)$ rather than necessarily on the raw solver state,
where $\mathcal D$ indexes the domain or model family. For ZINC,
$\chi_{\mathcal D}$ is the identity. For the Cora, Citeseer, and Pubmed
citation networks \cite{sen2008collective,yang2016revisiting}, CLUSTER
\cite{dwivedi2020benchmarking}, MNIST \cite{lecun1998gradient}, and CIFAR-10
\cite{krizhevsky2009cifar}, $\chi_{\mathcal D}(z)=\tanh(z)$, as in the
evaluated models.

The chemistry experiments use the ZINC molecular-property benchmark
\cite{irwin2012zinc}. The node-classification experiments use the Cora,
Citeseer, and Pubmed citation graphs
\cite{sen2008collective,yang2016revisiting}. The long-range
node-classification experiments use the CLUSTER benchmark
\cite{dwivedi2020benchmarking}. These non-vision experiments use
\begin{align}
\tilde z &= \chi_{\mathcal D}(z),
\label{eq:interaction-signal}\\
f_\theta(z, x) &= \mathrm{LayerNorm}\Big(\mathrm{ReLU}\big(W_{\mathrm{stim}} x + L(\tilde z) + G(\tilde z)\big)\Big).
\label{eq:update}
\end{align}
Here $W_{\mathrm{stim}}x$, $L(\tilde z)$, and $G(\tilde z)$ all have the same
shape as $z$. The matrix $W_{\mathrm{stim}}$ maps the encoded input into the
state space. The normalization is standard layer normalization
\cite{ba2016layernorm}. The map $L$ is Graph Attention Network (GAT)-style attention
\cite{velickovic2018gat}
(Section~\ref{sec:local}), and $G$ is a mean-field broadcast over graph nodes
(Section~\ref{sec:global}). The vision experiments use MNIST
\cite{lecun1998gradient} and CIFAR-10 \cite{krizhevsky2009cifar}. They use a
distinct, simpler composition without the outer LayerNorm/ReLU,
\begin{align}
\tilde z &= \tanh(z),
\label{eq:vision-interaction-signal}\\
f_\theta(z,x) &= W_{\mathrm{stim}} x + L_{\mathrm{vis}}(\tilde z) + G_{\mathrm{vis}}(\tilde z),
\label{eq:update-vision}
\end{align}
where $L_{\mathrm{vis}}$ and $G_{\mathrm{vis}}$ are vision-specific rather than
instances of $L$/$G$ above,
\begin{align}
L_{\mathrm{vis}}(\tilde z)_i
&= \frac{1}{|\mathcal N(i)|}\sum_{j\in\mathcal N(i)} W_L \tilde z_j,
\label{eq:vision-local-term}\\
G_{\mathrm{vis}}(\tilde z)
&= \tilde z\,\mathrm{softmax}\!\big((W_q \tilde z)(W_k \tilde z)^\top/\sqrt d\big).
\label{eq:vision-global-term}
\end{align}
In Eq.~\eqref{eq:vision-local-term}, $i$ and $j$ index hidden-channel nodes
within one sample, $|\mathcal N(i)|$ is the number of selected neighbors, and
$W_L$ is shared across all such neighbors. In
Eq.~\eqref{eq:vision-global-term}, $W_q$ and $W_k$ are learned query and key
projections applied to $\tilde z$ for the channel self-attention map.
$L_{\mathrm{vis}}$ is a
degree-normalized average over a
$k$-nearest-neighbor graph in \emph{channel space}, meaning that
$\mathcal N(i)$ ranges over the $d$ hidden channels of one sample, not over
other samples or graph nodes. It uses one shared linear map $W_L$ and no
per-edge learned attention weights. $G_{\mathrm{vis}}$ is genuine
self-attention, but over that one sample's own $d$ channels, not a mean-field
vector pooled across nodes or samples. Both compositions are damped by the
same Eq.~\eqref{eq:picard}, so the outer fixed-point template is shared across
every domain. Because the exact local and global maps vary, the theoretical
results below (spectral radius, energy diagnostic, local descent surrogate, and
path-sum) are stated
generically in terms of $f_\theta$ and its Jacobian. They apply to either
composition. The graph-attention and mean-field-broadcast descriptions used in
the rest of this section refer specifically to Eq.~\eqref{eq:update},
not Eq.~\eqref{eq:update-vision}.

In the equations below, the coordinate $z_i$ denotes the state of node, atom,
channel, or hidden unit $i$, and an interaction coefficient indexed by $(i,j)$
describes influence from source $j$ to receiver $i$. The stimulus is
$s_\theta(x)=W_{\mathrm{stim}}x$, with any bias absorbed into the encoded input
or into $s_\theta$. The symbol $\tilde z$ denotes the recurrent signal consumed
by local and global maps. The symbols $L_\theta$ and $G_\theta$ denote the local
and global update maps executed inside the solver. Their local linearizations
at the evaluated state are $M_L$ and $M_G$, and $M=M_L+M_G$ denotes the
one-step interaction operator used only in the linearized analysis. The damped
step operators $P_{\mathrm{self}}$, $P_{\mathrm{local}}$, and
$P_{\mathrm{global}}$, the one-step propagator $\Pi_{i\leftarrow j}$, and the
response operator $R_{K,\alpha}$ are derived objects, not additional learned
parameters. Indices $i,j$ denote nodes or state coordinates, $k,t$ denote solver
iteration, and $r$ denotes path length or matrix power. Batched computations may
store states as row tensors, but the displayed equations use this
receiver-first coordinate convention.

\begin{figure}[H]
\centering
\includegraphics[width=1\textwidth]{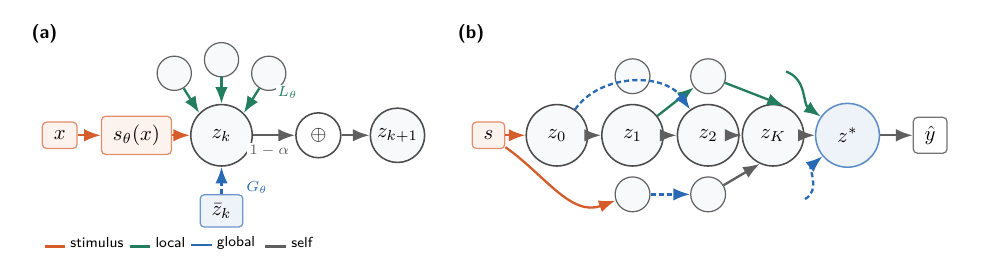}
\caption{Conceptual mechanism of a SILVA equilibrium layer. (a) Stimulus,
local interaction, global context, and self-persistence in the damped update of
Eqs.~\eqref{eq:picard}--\eqref{eq:interaction-step}. (b) Finite
stimulus-to-state paths and vector-attractor readout associated with
Eq.~\eqref{eq:source-self-local-global}.}
\label{fig:silvamechanism}
\end{figure}

\begin{definition}[SILVA layer]
A SILVA layer is a damped implicit layer with state
$z\in\mathbb R^{N\times d}$, stimulus $s_\theta(x)$, domain-specific
interaction maps $L_\theta$ and $G_\theta$, and dynamic interaction field
\begin{align}
\mathcal F_\theta(z,x) &= f_\theta(z,x)-z,
\label{eq:interaction-field}\\
z_{k+1} &= z_k+\alpha\,\mathcal F_\theta(z_k,x).
\label{eq:interaction-step}
\end{align}
A \emph{vector attractor} is a locally stable zero of this field,
$\mathcal F_\theta(z^\ast,x)=0$, equivalently
$z^\ast=f_\theta(z^\ast,x)$. Under the local linearization used for the
path-sum diagnostics, writing $s=W_{\mathrm{stim}}x$ and
$M=M_L+M_G$ gives
\begin{equation}
z_{k+1}\approx \alpha s
+\big((1-\alpha)I+\alpha M_L+\alpha M_G\big)z_k.
\label{eq:source-self-local-global}
\end{equation}
Here $I$ is the identity operator on the flattened $Nd$-dimensional state,
$s\in\mathbb R^{N\times d}$ is the stimulus reshaped as a state-sized tensor,
and $M_L,M_G\in\mathbb R^{Nd\times Nd}$ are the local and global Jacobian-level
operators at the evaluated state.
The finite solver therefore propagates an external stimulus through three
operator types. Self-persistence contributes $(1-\alpha)I$, local interactions
contribute $\alpha M_L$, and global interactions contribute $\alpha M_G$.
The layer uses damped identity persistence for the self term, with local and
global maps providing the learned or state-dependent interactions.
Figure~\ref{fig:silvamechanism}a gives the graphical reading of
Eqs.~\eqref{eq:picard}--\eqref{eq:interaction-step}. The stimulus arrow is
$s_\theta(x)$, the local and global arrows are $L_\theta(\tilde z_k)$ and
$G_\theta(\tilde z_k)$, and the gray arrow is the damped self-persistence
contribution $(1-\alpha)z_k$. The combination node represents the formation of
the next solver iterate $z_{k+1}$. Figure~\ref{fig:silvamechanism}b unfolds the
same update across solver time. Colored stimulus-to-state paths visualize the
finite powers in Eq.~\eqref{eq:source-self-local-global}, and the final state is
the approximate vector attractor read out by the task head. The diagram links
stimulus, interaction operators, damping, finite paths, and attractor readout
using the same notation as the equations.
\end{definition}

\paragraph{Linear equilibrium limit.}
The simplest limit of the SILVA update makes the stimulus-response structure
explicit. Let $b\in\mathbb R^N$ be a state vector, let
$m=s_\theta(x)=W_{\mathrm{stim}}x+c$ be the stimulus, and let
$A\in\mathbb R^{N\times N}$ be a receiver-first interaction matrix whose entry
$A_{ij}$ carries influence from source coordinate $j$ to receiver coordinate
$i$. The linear equilibrium condition is
\begin{equation}
b=m+Ab.
\label{eq:linear-bridge}
\end{equation}
Rearranging gives $(I-A)b=m$ and therefore
\begin{equation}
b=(I-A)^{-1}m,
\label{eq:linear-bridge-solution}
\end{equation}
whenever $I-A$ is invertible. If $\rho(A)<1$, the inverse is the Neumann
series $(I-A)^{-1}=\sum_{t=0}^{\infty}A^t$, so the hidden state is the sum of
the direct stimulus plus all repeated intra-layer interaction paths. SILVA
extends this linear limit by replacing the single matrix $A$ with the locally
linearized Jacobian of a domain-specific nonlinear update map whose terms are
stimulus, local interaction, and global interaction, with the nonlinearities and
normalization stated in
Eqs.~\eqref{eq:update}--\eqref{eq:update-vision}. The state is reached by the
damped solver of Eq.~\eqref{eq:picard} rather than by forming an explicit
inverse.

By the Banach fixed-point theorem \cite{banach1922operations}, a unique fixed
point exists and damped iteration converges whenever $f_\theta(\cdot,x)$ is a
contraction. In the locally linearized analysis used for the diagnostic, the
corresponding fixed-point stability condition is
\begin{align}
\rho\big(J_f(z^\ast)\big) &< 1,
\label{eq:contraction}\\
J_f(z^\ast) &= \partial f_\theta/\partial z\,\big|_{z^\ast},
\label{eq:fixed-point-jacobian}
\end{align}
where $\rho$ is the spectral radius of the Jacobian at the fixed point. The
Jacobian $J_f(z^\ast)\in\mathbb R^{Nd\times Nd}$ is taken with respect to the
flattened state coordinates of $z$, with $x$ held fixed. The damped solver
itself has local iteration Jacobian
\begin{equation}
J_{\mathcal T_\alpha}(z^\ast)=(1-\alpha)I+\alpha J_f(z^\ast).
\label{eq:damped-jacobian}
\end{equation}
The spectral-radius diagnostic estimates
$\rho_{\mathcal T}=\rho(J_{\mathcal T_\alpha})$, the spectral radius of the
actual damped update used by the finite solver. This diagnostic coincides with
$\rho(J_f)$ when $\alpha=1$, but otherwise measures the local spectral
stability of the executed iteration rather than the undamped update alone.
Stacked
SILVA architectures use multiple equilibria. For a stack of $L$ SILVA layers,
layer $\ell$ has its own parameters $\theta_\ell$,
state $z_\ell$, damping rate $\alpha_\ell$, solver budget $K_\ell$, update
map $f_{\theta_\ell}^{(\ell)}$, and connector $\psi_\ell$ to the next layer.
\begin{align}
z_{\ell,0} &= 0,
\label{eq:stacked-init}\\
z_{\ell,k+1} &= (1-\alpha_\ell)z_{\ell,k}
  +\alpha_\ell f_{\theta_\ell}^{(\ell)}(z_{\ell,k},x_\ell),
\label{eq:stacked-silva}\\
x_{\ell+1} &= \psi_\ell(z_{\ell,K_\ell}).
\label{eq:stacked-connector}
\end{align}
Here $\ell$ indexes equilibrium layers, $z_{\ell,k}$ is the $k$th solver iterate
inside layer $\ell$, $x_\ell$ is the input supplied to that layer, and
$\psi_\ell$ is a learned or fixed connector that maps the final state of layer
$\ell$ into the input space of layer $\ell+1$.
Stacking is therefore a composition of separately solved vector attractors,
not an increase in Picard steps inside one solver. The first attractor receives
the encoded data. Later attractors receive transformed states from earlier
attractors and solve their own local--global interaction fields. This
construction is related to multiscale DEQs \cite{bai2020mdeq}, but differs in
the equilibrium that is solved. MDEQ solves one enlarged equilibrium
$Z^\ast=F_\Theta(Z^\ast,x)$ in which several feature resolutions are
coupled and driven to equilibrium simultaneously. The sequential SILVA stack
solves each stage as its own root problem, with distinct
parameters and a distinct timescale, before its vector attractor is passed
to the next stage. In a local linearization, MDEQ has one coupled response
operator over the joint multiscale state, whereas a SILVA stack composes
stagewise responses,
$R_{\mathrm{stack}}\approx R_LD\psi_{L-1}\cdots R_2D\psi_1R_1$, with
$R_\ell$ the finite- or infinite-iteration response of layer $\ell$ and
$D\psi_\ell$ the Jacobian of the connector $\psi_\ell$ at the passed state.
Interaction histories are first accumulated within one stimulus-driven
attractor and then handed to the next, instead of all scales being solved as
one simultaneous equilibrium. In the main two-layer stack used for CIFAR-10,
ZINC, citation networks, and CLUSTER, Layer~1 ($\alpha_1=0.5$, fast)
feeds $\tanh(z_1^\ast)$ to Layer~2 ($\alpha_2=0.2$, slow). CIFAR-10 and ZINC
use $K_1=K_2=20$ Picard iterations. The node-classification and graph-benchmark
experiments use $K_\ell=15$ unless a specific ablation changes the stack. The
MNIST dynamic-channel-graph experiments use a single equilibrium layer with $\alpha=0.25$
and $K=20$. More generally, any number of such equilibrium layers can be
stacked, each at its own $\alpha_\ell$, giving a hierarchy of timescales analogous in spirit to
clockwork and hierarchical recurrent architectures
\cite{koutnik2014clockwork,elhihi1995hierarchical}. Here, every rung is a full
equilibrium solve over the interaction structure below, not a fixed-depth
recurrent step. The stacked-equilibrium result below evaluates a three-rung instance.

\paragraph{Relation to classical attractor and interaction models.}
Several established models provide useful reference points for this
formulation. Hopfield networks introduced recurrent collective computation as
convergence to associative memory states, usually under symmetry assumptions
that support an energy interpretation \cite{hopfield1982neural}. Wilson-Cowan
and Amari models describe population dynamics and continuous neural fields with
biologically interpretable activity variables and connection kernels
\cite{wilson1972excitatory,amari1977dynamics}. Predictive-coding and
feedback accounts of visual processing emphasize interactions between sensory
drive and contextual or top-down signals
\cite{rao1999predictive,gilbert2013topdown}. Interaction networks and
message-passing networks learn relations among objects or graph nodes
\cite{battaglia2016interaction,gilmer2017neural}. These lines of work motivate
the language of recurrent interaction, but the model studied here is a learned
implicit neural layer rather than a biological circuit or a symmetric
associative memory. Its attractor is the fixed point of the learned update map
reached by damped numerical iteration. Its local and global operators are
domain-specific neural modules trained from data. The additional structure is
the explicit separation of stimulus, self-persistence, local interactions, and
global interactions inside the solver, which makes the resulting finite
path-response analysis possible.

\paragraph{Evaluated model families.}
The same damped outer iteration in Eq.~\eqref{eq:picard} is used throughout,
but the state space, local graph, global term, nonlinearity, and backward mode
are domain-specific. In the vision experiments on MNIST and CIFAR-10, state
nodes are the hidden channels of one sample. The local graph is a dynamic
channel $k$-nearest-neighbor graph. The local term is a degree-normalized
neighbor average through a shared matrix $W_L$, the global term is per-sample
channel self-attention, and both consume $\tanh(z)$. The update map is a raw
sum. The finite-solver vision sweeps use Picard iteration with truncated
backpropagation. Anderson acceleration
\cite{anderson1965iterative,walker2011anderson} appears only in solver comparisons, and
the separate adjoint-trained MNIST diagnostic uses an undamped two-layer
interacting equilibrium model with a hidden-channel nearest-neighbor attention
local branch, a learned static global interaction matrix, and a generalized
minimal residual (GMRES) adjoint
\cite{saad1986gmres}. This diagnostic is used for the hidden-state geometry and
stability analyses rather than as the principal MNIST accuracy model.

For citation networks, state nodes are graph nodes and the local graph is the
observed citation edge set. The local term is edge-based graph attention on
$\tanh(z)$. The global term is a scalar-gated mean-field broadcast on
$\tanh(z)$, with bounded top-$k$ attention used only in the optional global
variant. For CLUSTER, state nodes are benchmark graph nodes, the local graph is
the benchmark edge set, the local term is graph attention on
$\tanh(z)$, and the global term is a mean-field broadcast on $\tanh(z)$ unless
it is explicitly ablated. These graph families use the
LayerNorm-ReLU update block, Picard iteration, and truncated backpropagation
through the unrolled solver. For ZINC, state nodes are atoms, the local graph
is the molecular bond graph, and the enhanced model includes bond attributes.
Both local and global terms consume raw $z$. The local term is graph attention
or bond-aware graph attention, while the global term is graph mean pooling
followed by a learned broadcast. ZINC also uses the LayerNorm-ReLU update
block, Picard iteration, and truncated backpropagation.

\begin{algorithm}[h]
\caption{Forward pass for one damped local--global equilibrium layer}\label{alg:forward-layer}
\begin{algorithmic}[1]
\State Encode raw input as $x$ and initialize $z_0$ (zeros or a learned/input-derived initial state)
\For{$k=0,\dots,K-1$}
  \State Compute the stimulus $s \leftarrow W_{\mathrm{stim}}x$
  \State Compute the recurrent signal $\tilde z_k\leftarrow\chi_{\mathcal D}(z_k)$
  \State Build or read the local graph $\mathcal N^{(k)}$ for this domain
  \State Compute $h_{\mathrm{local}}\leftarrow L(\tilde z_k,\mathcal N^{(k)})$
  \State Compute $h_{\mathrm{global}}\leftarrow G(\tilde z_k)$
  \State Form $f_\theta(z_k,x)$ using the domain-specific composition above
  \State Update $z_{k+1}\leftarrow(1-\alpha)z_k+\alpha f_\theta(z_k,x)$, or replace this step by the Anderson extrapolation of Eq.~\eqref{eq:anderson}
\EndFor
\State Return $z_K$ as the approximate fixed point and apply the task readout $W_{\mathrm{out}}z_K$
\end{algorithmic}
\end{algorithm}
Algorithm~\ref{alg:forward-layer} uses the same notation as
Eq.~\eqref{eq:picard}. The neighborhood object $\mathcal N^{(k)}$ is either a
fixed graph, as in molecules and citation networks, or a state-dependent graph,
as in the vision channel $k$-nearest-neighbor construction. The final matrix $W_{\mathrm{out}}$
denotes the task readout introduced in the Training subsection.

\subsection{Interaction-operator decomposition}\label{sec:grand}

Linearizing Eq.~\eqref{eq:picard} at fixed attention/gating weights gives a
block-structured operator $\Omega$ on the flattened state, decomposed
into three mechanistically distinct components,
\begin{equation}
\Omega = \Omega_{\mathrm{self}} + \Omega_{\mathrm{local}} + \Omega_{\mathrm{global}},
\label{eq:omega}
\end{equation}
Here $\Omega$ is the Jacobian-level interaction operator, not a separate
stored weight matrix. The three terms are $\Omega_{\mathrm{self}}$ (diagonal,
node-wise persistence), $\Omega_{\mathrm{local}}$ (sparse, neighborhood
message passing), and $\Omega_{\mathrm{global}}$ (dense low-rank, mean-field
context). Node persistence is provided by the
$(1-\alpha)z_k$ term of the damped iteration (Eq.~\eqref{eq:picard}) rather
than a separate learned self-weight, so the model comprises two
directly-parameterized interaction components (local, global) plus this
implicit self-term.

\paragraph{Solver-level operator interpretation.} These operators are more than
diagnostic labels. They correspond to the neural computations executed at each
solver step. In the locally linearized form used for the
diagnostics, one damped step can be written
\begin{align}
z_{k+1} &= \alpha s+\big(P_{\mathrm{self}}+P_{\mathrm{local}}+P_{\mathrm{global}}\big)z_k,
\label{eq:solver-operator-split}\\
P_{\mathrm{self}} &= (1-\alpha)I,
\label{eq:self-operator}\\
P_{\mathrm{local}} &= \alpha M_L,
\label{eq:local-operator}\\
P_{\mathrm{global}} &= \alpha M_G,
\label{eq:global-operator}
\end{align}
where $s=W_{\mathrm{stim}}x$ is the injected stimulus. The matrices $M_L$
and $M_G$ are the local linearizations of the evaluated local and global
branches. They include the map $z\mapsto\tilde z$ and any active
normalization/nonlinearity factors.
The operators
$P_{\mathrm{self}}$, $P_{\mathrm{local}}$, and $P_{\mathrm{global}}$ all act on
the same flattened state space as $M_L$ and $M_G$. Starting from
$z_0=0$,
\begin{align}
z_K
&=\alpha\sum_{t=0}^{K-1}
\big(P_{\mathrm{self}}+P_{\mathrm{local}}+P_{\mathrm{global}}\big)^t s,
\label{eq:solver-path-operator}\\
z_K
&=\alpha\sum_{t=0}^{K-1}
\sum_{w\in\{\mathrm{self},\mathrm{local},\mathrm{global}\}^t}
P_{w_t}\cdots P_{w_1}s .
\label{eq:solver-path-words}
\end{align}
Each word $w=(w_1,\ldots,w_t)$ is a concrete stimulus-to-state interaction
history through the evaluated layer. A self symbol keeps information through
damping, a local symbol applies the neighborhood update, and a
global symbol applies the broadcast or attention-mediated context update. The
coordinate $(P_{w_t}\cdots P_{w_1}s)_i$ is the contribution of
that particular interaction history to node or channel $i$. The later
Neumann-series path-sum form (Section~\ref{sec:pathsum}) is the same expansion after
collecting the self-persistence factors and writing the remaining interaction
operator as $M=M_L+M_G$. For $t=0$, the word is empty and the product is the
identity, giving the direct stimulus contribution.

Written per-channel for the undamped map $f_\theta$, the update for feature
channel $c$ of node $i$ is
\begin{equation}
f_\theta(z^{(k)},x)_{i,c} = \sigma\Bigg(
\underbrace{u_{i,c}}_{\text{stimulus}} +
\underbrace{\sum_{j\in\mathcal{N}(i)}\alpha_{ij}^{(k)}\sum_f W_{cf}^{\mathrm{local}} \tilde z_{j,f}^{(k)}}_{\text{local, }\Omega_{\mathrm{local}}} +
\underbrace{\tfrac1N\sum_j\sum_f W_{cf}^{\mathrm{global}} \tilde z_{j,f}^{(k)}}_{\text{global, }\Omega_{\mathrm{global}}}
\Bigg).
\label{eq:scalar-master}
\end{equation}
In Eq.~\eqref{eq:scalar-master}, $u=s_\theta(x)$ is the stimulus, $i$ is the receiver node, $j$ is a source
node, $c$ is the receiver feature channel, and $f$ is the source feature
channel being summed through the channel-mixing matrices. The coefficient
$\alpha_{ij}^{(k)}$ is either a learned attention weight or a normalized edge
weight at solver iteration $k$, depending on the domain-specific local
operator. The recurrent signal is
$\tilde z^{(k)}=\chi_{\mathcal D}(z^{(k)})$, and $\sigma$ denotes the domain's
outer normalization/nonlinearity block or the identity map when no such block
is used. The global sum uses all $N$ nodes in the
same sample, molecule, or graph. The matrices
$W_{\mathrm{local}},W_{\mathrm{global}}\in\mathbb R^{d\times d}$ mix feature
channels, while $\sigma$ acts on the resulting node-feature vector. The damped
solver then combines this map with the previous state according to
Eq.~\eqref{eq:picard}.

\paragraph{Unified local--global interaction sum.} The split does not introduce two
unrelated design pieces. It can be written as one per-node interaction sum
whose sparse local channel and dense global channel have different support.
Define a one-step propagator $\Pi_{i\leftarrow j}$ describing how source node
$j$ contributes to receiver node $i$,
\begin{equation}
\Pi_{i\leftarrow j}(\tilde z_j) =
\mathbf{1}_{\{j\in\mathcal N(i)\}}\alpha_{ij}\,W_{\mathrm{local}}\,\tilde z_j
+\tfrac1N\,W_{\mathrm{global}}\,\tilde z_j,
\label{eq:propagator}
\end{equation}
where $\Pi_{i\leftarrow j}$ maps the source feature vector $\tilde z_j$ to a
contribution at receiver $i$. The indicator selects the local neighborhood
term and the second term is the mean-field contribution from every node. The
matrices $W_{\mathrm{local}}$ and $W_{\mathrm{global}}$ mix feature channels.
The update can then be
written as a single sum over \emph{all} $N$ nodes,
$f_\theta(z^{(k)},x)_i =
\sigma\big(u_i + \sum_{j=1}^N \Pi_{i\leftarrow j}(\tilde z_j^{(k)})\big)$.
The local/global split is therefore a support and operator choice within one
interaction sum. Collecting the same contributions at the level of the full
state vector gives the linear operator's closed form directly, using the
adjacency matrix $A$ ($A_{ij}=\alpha_{ij}$ for $j\in\mathcal N(i)$, else $0$)
and the Kronecker product $\otimes$ to carry the per-node channel mixing
$W_\bullet$ across all $N$ nodes simultaneously.
\begin{equation}
J_f(z^\ast) \,\approx\,
\underbrace{A \otimes W_{\mathrm{local}}}_{\Omega_{\mathrm{local}}}
\,+\, \underbrace{\tfrac1N\,\mathbf{1}\mathbf{1}^\top \otimes W_{\mathrm{global}}}_{\Omega_{\mathrm{global}}},
\label{eq:omega-kron}
\end{equation}
Equation~\eqref{eq:omega-kron} is Eq.~\eqref{eq:omega} written out as an
explicit $Nd\times Nd$ matrix. The first term is sparse with the local graph's
sparsity pattern. The second term is rank-1 dense in node space. Since
$\mathbf{1}\mathbf{1}^\top$ is the all-ones matrix, this term repeats the
same $W_{\mathrm{global}}$ block at every $(i,j)$ pair, which is exactly what
mean-field means at the matrix level. The expression is a linearization at
fixed attention/gating weights $\alpha_{ij}$, frozen at their currently
evaluated value. It does not make
$\mathrm{LayerNorm}(\mathrm{ReLU}(\cdot))$ in Eq.~\eqref{eq:update} itself
linear.

\subsection{Local interaction ($\Omega_{\mathrm{local}}$)}\label{sec:local}

For chemistry and graph-node experiments, let $y$ denote the signal supplied to
the local map. Inside the solver, $y=\tilde z=\chi_{\mathcal D}(z)$. The local
term follows GAT-style attention \cite{velickovic2018gat},
\begin{align}
L(y)_i &= \sum_{j\in\mathcal{N}(i)} a_{ij} W y_j,
\label{eq:local-gat}\\
a_{ij} &= \mathrm{softmax}_j\!\left(\mathrm{LeakyReLU}(u^\top[Wy_i\|Wy_j])\right),
\label{eq:local-gat-attention}
\end{align}
The neighborhood $\mathcal{N}$ is the
molecular bond graph for chemistry or the citation graph for node
classification, an instance of the general neural message-passing framework
\cite{gilmer2017neural}, of which graph convolutional networks
\cite{kipf2017gcn} are the special case with fixed, non-attentive edge
weights. In this local-attention formula, $W$ maps node features into the
attention feature space, $u$ is the learned attention vector, $\|$ denotes
feature concatenation, and the softmax is normalized over source nodes
$j\in\mathcal N(i)$ for each receiver $i$. Vision's local term, $L_{\mathrm{vis}}$
(Eq.~\eqref{eq:update-vision}), is architecturally different. It changes both
the neighborhood construction and the update formula. It is a
degree-normalized average through one shared linear map, with no per-edge
attention weight $a_{ij}$ at all, over a symmetrized $k$-nearest-neighbor
graph built in \emph{channel space}. The neighbors of channel $i$ are other
hidden channels of the same sample, not other samples or graph nodes.

\subsection{Domain-specific input encoders}

Raw input reaches the equilibrium layer through a domain-specific encoder,
$x = \mathcal{R}_\phi(x_{\mathrm{raw}})$. For MNIST the encoder is a single learned linear
projection,
\begin{align}
x &= W_{\mathrm{proj}}\,x_{\mathrm{raw}},
\label{eq:mnist-projection}\\
x_{\mathrm{raw}} &\in\mathbb{R}^{784},
\label{eq:mnist-input-space}\\
W_{\mathrm{proj}} &\in\mathbb{R}^{d\times 784}.
\label{eq:mnist-projection-space}
\end{align}
Here $x_{\mathrm{raw}}$ is the flattened $28\times28$ image, $W_{\mathrm{proj}}$
is the learned encoder matrix, and $x\in\mathbb R^d$ is the hidden vector passed
to the first equilibrium layer.
For CIFAR-10 the encoder is a two-block convolutional stack with batch normalization,
max-pooling, and dropout \cite{ioffe2015batchnorm,srivastava2014dropout}.
For $i=1,2$,
\begin{align}
h_i &= \mathrm{MaxPool}_2\big(\mathrm{ReLU}(\mathrm{BN}_i(\mathrm{Conv}_i(h_{i-1})))\big),
\label{eq:cifar-conv-block}\\
h_0 &= x_{\mathrm{raw}}.
\label{eq:cifar-input-state}
\end{align}
\begin{equation}
x = W_{\mathrm{proj}}\,\mathrm{Flatten}\big(\mathrm{Dropout}_{p=0.3}(h_2)\big),
\label{eq:cifar-projection}
\end{equation}
Here $h_i$ is the feature map after convolution block $i$,
$\mathrm{Conv}_i$ and $\mathrm{BN}_i$ are the convolution and batch
normalization in that block, and $\mathrm{MaxPool}_2$ denotes $2\times2$
max-pooling. The matrix $W_{\mathrm{proj}}$ maps the flattened post-dropout
feature map to the hidden vector $x\in\mathbb R^d$. The stack uses
$3\!\to\!32\!\to\!64$ output channels, $3\times3$ kernels with padding
1, and dropout applied only during training. The encoder is a shallow
feature extractor that hands its representation to the equilibrium dynamics.

\subsection{Domain-specific interaction graphs}

The architecture's motivating premise is that vision and chemistry can share
the same damped local+global equilibrium template even though neither the
node set nor the local mechanism is identical between them. Chemistry's
nodes are atoms with GAT-style local attention over the fixed bond graph
(Section~\ref{sec:local}), while vision's ``nodes,'' for the purpose of its
own local term $L_{\mathrm{vis}}$, are the $d$ hidden channels of one
sample, with a degree-normalized average (not attention) over
\begin{equation}
\mathcal{N}^{(k)}(i) = \operatorname*{arg\,top\text{-}k}_{j\neq i}
\big(-\|\tilde z_i^{(k)} - \tilde z_j^{(k)}\|_2\big),
\label{eq:vision-knn}
\end{equation}
where $\tilde z^{(k)}=\tanh(z^{(k)})$ for the vision models. The
neighborhood is recomputed from the current recurrent signal, not from
$z^\ast$, at every solver iteration. Equation~\eqref{eq:vision-knn} is dynamic
and content-dependent, uses Euclidean distance over the current hidden state's
channels, and is symmetrized
(Section~\ref{sec:local} above). The operator $\operatorname*{arg\,top\text{-}k}$
returns the set of $k$ source-channel indices with largest score, equivalently
the nearest neighbors because the score is the negative distance. The construction
resembles dynamic graph CNNs, which recompute feature-space neighborhoods
inside the network, but the entities here are hidden channels rather than
points in a point cloud \cite{wang2019dgcnn}. For chemistry the
neighborhood graph $\mathcal{N}(i)$ is the molecule's fixed bond graph,
identical at every iteration. The shared object is the damped local+global
equilibrium template, while the graph construction is domain-specific. Vision
infers a dynamic hidden-channel graph from recurrent state similarity, and
chemistry uses the fixed molecular bond graph.

\subsection{Global interaction ($\Omega_{\mathrm{global}}$)}\label{sec:global}

For chemistry and graph-node experiments, the global term computes a
mean-field vector within one system (one molecule's $N$ atoms or one
graph's $N$ nodes) and broadcasts it back to that system's own nodes. For a
single system with interaction signal $y$,
\begin{equation}
G(y) = \tfrac{1}{N}\mathbf{1}_N\mathbf{1}_N^\top y W_g^\top,
\label{eq:mean-field-global}
\end{equation}
where $y=\tilde z$ inside the solver and $W_g$ is the learned channel-mixing
matrix. The vector $\mathbf{1}_N$ is the all-ones vector in $\mathbb R^N$, so
$N^{-1}\mathbf{1}_N\mathbf{1}_N^\top y$ replaces
each row by the graph or molecule mean state. Equation~\eqref{eq:mean-field-global}
is a fixed averaging operator over nodes followed by learned channel mixing.
Vision's
global term, $G_{\mathrm{vis}}$ (Eq.~\eqref{eq:update-vision}), is not an
instance of this mean-field broadcast. It is genuine self-attention, but
over one sample's own $d$ hidden channels rather than pooled across
samples or nodes, so there is no $N$-indexed pooling step in $G_{\mathrm{vis}}$
at all. For node
classification on real graphs, this reduces to an $O(N)$ scalar-gated
mean-field broadcast. The graph-wide mean state $g=\tfrac1N\sum_j y_j$ is
computed once, gated by a learned, data-dependent scalar and redistributed
to every node,
\begin{align}
\beta &= \sigma\!\left(\frac{(W_q g)^\top (W_k g)}{\sqrt{d}}\right),
\label{eq:global-gate}\\
G(y)_i &= \beta\,W_g\,g.
\label{eq:global-broadcast}
\end{align}
This broadcast is applied to every node $i$. Here $g\in\mathbb R^d$ is the
mean state, $\beta$ is a scalar gate, $d$ is the hidden dimension, and
$W_q, W_k, W_g$ are learned matrices distinct from the local term's own
parameters. The sigmoid $\sigma$ maps the scalar attention score to
$\beta\in(0,1)$. The gate uses the same bilinear,
$\sqrt{d}$-scaled construction as standard dot-product attention
(Eq.~\eqref{eq:update}'s own attention mechanism), here applied once to the
single graph-wide vector $g$ instead of pairwise between nodes. This gives an
$O(N)$ global term at graph scale \cite{vaswani2017attention}. Full pairwise
attention over a real citation graph produced an exactly uniform attention
pattern in the citation-network diagnostic below. A static
(non-gated) variant fixes
$\beta=1$ in Eq.~\eqref{eq:global-gate}. A bounded top-$k$ variant replaces
the single-vector gate of Eq.~\eqref{eq:global-gate} with genuine
node-to-node dot-product attention, restricted to each node's $k$ nearest
candidates (with $O(N^2)$ score computation, avoiding the uniform-attention
collapse noted above since the softmax is always taken over only $k$
candidates regardless of $N$),
\begin{align}
G(y)_i &=
\sum_{j\in\mathcal{T}_k(i)}
\mathrm{softmax}_j\!\left(\frac{(W_q y_i)^\top (W_k y_j)}{\sqrt{d}}\right) W_v y_j,
\label{eq:global-topk}\\
\mathcal{T}_k(i)
&= \operatorname*{arg\,top\text{-}k}_{j} (W_q y_i)^\top(W_k y_j),
\label{eq:global-topk-set}
\end{align}
recomputed from the current interaction signal every solver iteration, with
$\mathcal T_k(i)$ the selected source-node set for receiver $i$ and
$W_q, W_k, W_v$ a third, independent set of learned matrices for this
variant. The softmax in Eq.~\eqref{eq:global-topk} is normalized only over
sources $j\in\mathcal T_k(i)$. The index $i$ is the receiving node, $j$ is a
candidate source node, and $W_v$ maps the selected source state into the value
vector that is summed. These variants define the global-term alternatives used
in the citation-network ablations below.

\subsection{Spectral-radius diagnostic at the fixed point}\label{sec:rho}

The damped-step spectral radius
$\rho_{\mathcal T}=\rho(J_{\mathcal T_\alpha}(z^\ast))$ is estimated by power iteration on
the vector--Jacobian product of the executed solver update
(Algorithm~\ref{alg:rho}). For a normal linearization, values below one
indicate local one-step contraction of the damped iteration. More generally,
because the Jacobian can be non-normal, $\rho_{\mathcal T}$ is used here as a
local asymptotic stability indicator rather than as a singular-value
contraction certificate. It is related to the Jacobian quantity whose crossing
of 1 explains vanishing or exploding gradients in ordinary recurrent networks
\cite{pascanu2013difficulty}, here tracked on the equilibrium itself instead
of on an unrolled training trajectory. In DEQ models, growth of the equilibrium
Jacobian is also known to affect both forward and backward solver stability
\cite{bai2021jacobian}. A Lyapunov exponent would require trajectory averaging
of $\log\|J\|$, with $\log\rho_{\mathcal T}$ serving only as a local
linearized proxy under the estimator used here.

\begin{algorithm}[h]
\caption{Spectral radius estimation at the fixed point}\label{alg:rho}
\begin{algorithmic}[1]
\State $v_0 \sim \mathcal{N}(0,I)$, $\|v_0\|=1$
\For{$t=1,\dots,T$}
  \State $v_t \leftarrow J_{\mathcal T_\alpha}(z^\ast)^\top v_{t-1}$ via the vector--Jacobian product (VJP) of the damped update
  \State $\rho \leftarrow \|v_t\|$
  \State $v_t \leftarrow v_t/\rho$
\EndFor
\end{algorithmic}
\end{algorithm}
In Algorithm~\ref{alg:rho}, $v_t$ is a unit vector in the flattened state
space, $T$ is the number of power-iteration steps, and VJP denotes a
vector--Jacobian product evaluated at the fixed point $z^\ast$. The scalar
$\rho$ returned after normalization is a power-iteration estimate of the
dominant eigenvalue magnitude of $J_{\mathcal T_\alpha}(z^\ast)$ under the
usual spectral assumptions for that estimator.

\subsection{Energy diagnostic and local descent surrogate}\label{sec:energy}

A companion diagnostic to $\rho$ is a scalar quantity tracking alignment between
the activated state and the global interaction acting on it, in the tradition of
Hopfield-style energy functions for recurrent networks
\cite{hopfield1982neural} and the broader energy-based view of learning
\cite{lecun2006tutorial}. The MNIST solver-dynamics analysis uses the
evaluated diagnostic
\begin{align}
a_b(z) &= \tanh(z_b),
\label{eq:energy-activation}\\
E_{\mathrm{diag}}(z)
&= \frac{1}{B}\sum_{b=1}^{B}
\left[a_b(z)\cdot a_b(z) -
a_b(z)\cdot\big(a_b(z)A_{\mathrm{global}}^{(b)}\big)\right].
\label{eq:energy-diagnostic}
\end{align}
Here $b$ indexes samples in a batch of size $B$, $a_b(z)$ is the activated
hidden state for sample $b$, and $A_{\mathrm{global}}^{(b)}$ is the
row-normalized global-attention matrix computed for that sample at the current
solver iterate. This is the quantity plotted in Figure~\ref{fig:solverdynreal}.
It measures global-attention self-consistency in the vision model and
does not include the local term.

Treating a DEQ's fixed point as the equilibrium of a nonlinear dynamical
system and analyzing its stability directly is itself an active direction.
Chu et al.~\cite{chu2024lyapunov} enforce Lyapunov stability of DEQ fixed
points as an explicit training objective and show this improves adversarial
robustness. Here, the energy diagnostic and $\rho$ are evaluated on networks
trained without a stability constraint. The robustness-versus-spectral-radius analysis
evaluates whether observed variation in $\rho$ predicts robustness under input
perturbations, which is a different question from the stability-enforcing
objective of \cite{chu2024lyapunov}.

\paragraph{Local descent surrogate for the damped iteration.} The scalar
diagnostic plotted in Figure~\ref{fig:solverdynreal} and the quadratic
surrogate below serve distinct roles. For the
linearized interpretation of the damped Picard step, hold the update vector
$h=f_\theta(z_k,x)$ fixed at the current iterate and define
\begin{equation}
\mathcal E_h(z) = \tfrac12\|z\|^2 - z^\top h.
\label{eq:energy-surrogate}
\end{equation}
The state is interpreted as a flattened vector when the dot product is formed.
The surrogate is the local quadratic model obtained by freezing the nonlinear
update, the same local-linearization move used to define $\Omega$ in
Section~\ref{sec:grand}.
Its gradient is
\begin{equation}
\nabla_z \mathcal E_h(z) = z - h.
\label{eq:energy-grad}
\end{equation}
Setting $\nabla_z \mathcal E_h(z)=0$ recovers exactly the fixed-point condition
$z^\ast=h$, the frozen-update version of $z^\ast=f_\theta(z^\ast,x)$. A single
gradient-descent step on $\mathcal E_h$ with step size $\eta$,
\begin{align}
z_{k+1} &= z_k - \eta\,\nabla_z \mathcal E_h(z_k),
\label{eq:energy-descent}\\
z_{k+1} &= z_k - \eta(z_k - h),
\label{eq:energy-descent-gradient}\\
z_{k+1} &= (1-\eta)z_k + \eta\,h,
\label{eq:energy-descent-picard}
\end{align}
is, term for term, the damped Picard update of Eq.~\eqref{eq:picard} with
$\eta=\alpha$ and $h$ identified with $f_\theta(z_k,x)$. Under this local
surrogate, $\mathcal E_h$ is non-increasing under the usual gradient-descent
step-size condition for a locally quadratic objective. The global energy
function of the trained nonlinear network is not assumed.

\subsection{Fixed-point acceleration}

Under appropriate contractive assumptions, plain Picard iteration
(Eq.~\eqref{eq:picard}) converges linearly. Several experiments instead use
Anderson acceleration
\cite{anderson1965iterative,walker2011anderson}, the same fixed-point
extrapolation technique used in the original DEQ solver \cite{bai2019deq},
which extrapolates from the $m_{\mathrm A}$ most recent iterates by solving a
constrained least-squares problem for mixing coefficients $\gamma$ that
minimize the residual $r_k = f_\theta(z_k,x)-z_k$ in their span.
\begin{align}
\gamma
&= \operatorname*{arg\,min}_{\tilde\gamma}
\Big\|\textstyle\sum_{j=0}^{m_{\mathrm A}-1}\tilde\gamma_j r_{k-j}\Big\|_2,
\label{eq:anderson-coefficients}\\
\sum_j \tilde\gamma_j &= 1,
\label{eq:anderson-normalization}\\
z_{k+1}
&= \textstyle\sum_{j=0}^{m_{\mathrm A}-1}\gamma_j\, f_\theta(z_{k-m_{\mathrm A}+j+1}, x).
\label{eq:anderson}
\end{align}
Here $m_{\mathrm A}$ is the Anderson memory length, distinct from the stimulus
symbol $m$ used in the scalar examples. The vector $r_k$ is the residual of the
fixed-point equation at iterate $k$, $\tilde\gamma$ is the optimization variable
in the least-squares problem, and $\gamma_j$ are the resulting mixing weights
over the last $m_{\mathrm A}$ residuals. The constraint in
Eq.~\eqref{eq:anderson-normalization} makes the extrapolation affine, so the
next iterate is a weighted combination of recent update-map evaluations rather
than a newly parameterized layer.

\subsection{Path-sum (Neumann series) interpretation}\label{sec:pathsum}

The architecture's global-context design is motivated by a path-sum argument.
If $\Omega$ acted only once, indirect influence between nodes would be lost.
Summing every interaction path gives
$\Omega + \Omega^2 + \Omega^3 + \cdots = (I-\Omega)^{-1}\Omega$, converging
whenever $\rho(\Omega)<1$ as in Eq.~\eqref{eq:contraction}. The derivation then
connects this infinite-series picture to the finite-iteration solver,
since convergence alone does not say how much of the series a damped budget of
$K$ iterations resolves.

\paragraph{Three-node scalar path expansion.} Consider three nodes
$\{1,2,3\}$, each receiving a stimulus $m_i$ and interacting through an
elementary $3\times3$ matrix $A$ with entries $a_{ij}$ (the influence of
node $j$ on node $i$). The update rule $b = m + A\tanh(b)$ says node $i$
integrates its own stimulus $m_i$ plus a weighted sum of every other node's
nonlinear signal, $\sum_j a_{ij}\tanh(b_j)$, including a self-loop at
$j=i$. Linearizing ($\tanh(z)\approx z$ for small activations) makes the
path structure explicit. Starting from $b^{(0)}=m$ and iterating
$b^{(t+1)} \approx m + Ab^{(t)}$,
\begin{align}
b^{(1)} &\approx m + Am,
\label{eq:three-node-first-iterate}\\
b^{(2)} &\approx m + Am + A^2m,
\label{eq:three-node-second-iterate}\\
b^{(3)} &\approx m + Am + A^2m + A^3m.
\label{eq:three-node-third-iterate}
\end{align}
and by induction $b^{(T)} \approx \sum_{r=0}^{T} A^r m$. The $r$-step term
$(A^r m)_i$ sums every length-$r$ path into node $i$. Collecting all path
lengths gives the effective interaction
$A_{\mathrm{eff}} = A + A^2 + A^3 + \cdots = (I-A)^{-1}A$ whenever
$\rho(A)<1$. In the linear regime, $b^\ast \approx (I-A)^{-1}m$ exactly, so the
path interpretation is the closed form rather than an analogy. The same
statement holds for the locally linearized damped finite-$K$ solver, with $A$
generalized to $M = M_L + M_G$.

For the evaluated damped solver, the local linear form in
Eq.~\eqref{eq:source-self-local-global} can be written
\begin{align}
z_{t+1} &= \alpha s+T_\alpha z_t,
\label{eq:damped-path-operator}\\
T_\alpha &= (1-\alpha)I+\alpha M,
\label{eq:damped-transition}\\
M &= M_L+M_G.
\label{eq:local-global-linearization}
\end{align}
This gives $z_K=\alpha\sum_{t=0}^{K-1}T_\alpha^t s$. Expanding
$T_\alpha^t$ gives all stimulus-to-state paths made of self-persistence
steps and local/global interaction steps. Because the self-persistence term
is a scalar multiple of the identity, collecting those self steps yields the
weighted powers of $M$ below.

$M = J_f(z^\ast)$ decomposes at the local-linear level as $M = M_L + M_G$.
The factors in $M_L$ and $M_G$ include the derivative of
$z\mapsto\tilde z$ and any active outer normalization or nonlinearity, matching
$\Omega_{\mathrm{local}}+\Omega_{\mathrm{global}}$ of Eq.~\eqref{eq:omega}.
Solving the linearized
system $z=s+Mz$ ($s = W_{\mathrm{stim}}x$) by damped Picard iteration from
$z_0=0$ gives, in closed form,
\begin{align}
z_K &= \sum_{r=0}^{K-1} w_r^{(K,\alpha)} M^r s,
\label{eq:pathsum}\\
w_r^{(K,\alpha)}
&= \alpha\sum_{q=r}^{K-1}\binom{q}{r}(1-\alpha)^{q-r}\alpha^r.
\label{eq:path-weight}
\end{align}
Here $r$ is the interaction length, $q$ is the solver-step index being summed
out, and $w_r^{(K,\alpha)}$ is the finite-budget weight assigned to all
length-$r$ interaction products. Since $M=M_L+M_G$ and matrix products do not
commute, $M^r = \sum_{w\in\{L,G\}^r} M_{w_r}\cdots M_{w_1}$. It sums every
ordered length-$r$ path through the two-generator interaction algebra, where
path means an ordered product of local and global propagators acting on the
stimulus. Equation
\eqref{eq:pathsum} is an exact finite path-sum expansion with two
propagator types and external source $s$. As $K\to\infty$,
$w_r^{(K,\alpha)}\to1$ and Eq.~\eqref{eq:pathsum} recovers the Neumann series
$z^\ast=(I-M)^{-1}s=\sum_{r=0}^\infty M^r s$, independent of $\alpha$, which
only governs which partial sum is realized at finite $K$.

As $\rho(M)\to1^-$, terms $M^rs$ decay more slowly with $r$. Longer interaction
histories matter, and finite $K$ increasingly under-resolves them. This gives
the edge-of-chaos diagnostic (Figure~\ref{fig:edgeofchaosreal}) a precise
meaning, in the sense already established for recurrent and deep networks at
the boundary between ordered and chaotic dynamics
\cite{bertschinger2004real,poole2016exponential}. It tracks how close the
network sits to the boundary at which long-range interaction paths stop being
negligible.

\subsection{Linear response and effective influence}\label{sec:green}

Equation~\eqref{eq:pathsum} also defines the finite-iteration response
operator. In the linear-response sense, the Green's operator maps an injected
source or stimulus to the state produced by the locally linearized system.
\begin{align}
R_{K,\alpha} &= \sum_{r=0}^{K-1} w_r^{(K,\alpha)} M^r,
\label{eq:finite-green}\\
z_K &= R_{K,\alpha}s.
\label{eq:finite-response}
\end{align}
The finite solver budget realizes $R_{K,\alpha}$ as its Green's operator.
The indices $i$ and $j$ now index flattened state coordinates, which combine
node and feature-channel indices when $z\in\mathbb R^{N\times d}$. Its entry
$(R_{K,\alpha})_{ij}$ is the total linearized effect on state coordinate $i$
of a unit stimulus at coordinate $j$, after all interaction paths resolved by
$K$ damped iterations have been summed.
In the contractive infinite-iteration limit,
\begin{align}
R &= \lim_{K\to\infty}R_{K,\alpha},
\label{eq:green-limit}\\
R &= (I-M)^{-1},
\label{eq:green-resolvent}
\end{align}
independent of $\alpha$. This operator is an interpretability object attached
to the locally linearized trained layer, derived from the update map rather than
learned as an additional parameter. It is connected to the fitted layer through
the same update map and Jacobian-vector machinery used for
Algorithm~\ref{alg:rho}. The computation can use the linear action of $M$
through vector--Jacobian or Jacobian--vector products without materializing
$M$ or $R$.

\subsection{Training objectives}

Models are trained with negative log-likelihood loss for classification,
\begin{align}
\mathcal{L}_{\mathrm{cls}} &= -\log p_\theta(y \mid z^\ast),
\label{eq:classification-loss}\\
p_\theta(\cdot\mid z^\ast) &= \mathrm{softmax}(W_{\mathrm{out}} z^\ast),
\label{eq:classification-readout}
\end{align}
or mean absolute error for the ZINC regression target,
\begin{align}
\mathcal{L}_{\mathrm{reg}} &= \big|\, \hat y(z^\ast) - y \,\big|,
\label{eq:regression-loss}\\
\hat y(z^\ast) &= W_{\mathrm{out}} z^\ast,
\label{eq:regression-readout}
\end{align}
Here $\mathcal L_{\mathrm{cls}}$ is the per-example classification loss,
$p_\theta(y\mid z^\ast)$ is the predicted probability assigned to the true class
$y$, and $p_\theta(\cdot\mid z^\ast)$ is the full softmax distribution over
classes. The matrix $W_{\mathrm{out}}$ is the learned final readout. It maps a
node state, pooled graph state, or image hidden state to logits for
classification, and maps the pooled molecular state to a scalar prediction
$\hat y$ for regression. In Eq.~\eqref{eq:regression-loss}, $y$ is the
ground-truth scalar property and $\mathcal L_{\mathrm{reg}}$ is the absolute
error for one molecule. For node classification, the readout is applied row-wise
to node states. For graph-level classification or regression, $z^\ast$ is pooled
before the final readout. For image classification, $z^\ast$ is the final hidden
representation returned by the vision equilibrium stack. Optimization uses Adam \cite{kingma2015adam}, gradient clipping
\cite{pascanu2013difficulty}, and step-decay or plateau scheduling. The losses
contain no explicit physics term or fixed-point residual term. The dynamical
quantities above are diagnostics evaluated on trained networks.

\subsection{Backward differentiation}\label{sec:backward}

SILVA models are differentiated in two ways in these experiments. Most
vision, chemistry, node-classification, and graph-benchmark models differentiate
through the finite $K$ solver iterations. The separate adjoint-trained MNIST
diagnostic uses a GMRES linear solve over the adjoint operator described in the
Supplementary Methods. The two backward modes share the same
fixed-point motivation, but they correspond to different numerical training
procedures and are reported with different scope.

\paragraph{Finite unrolled backward.}
Let
\begin{equation}
\mathcal T_{\alpha,\theta}(z,x)=(1-\alpha)z+\alpha f_\theta(z,x)
\label{eq:damped-map-backward}
\end{equation}
be the damped solver map and let
$z_{k+1}=\mathcal T_{\alpha,\theta}(z_k,x)$ for
$k=0,\ldots,K-1$. The loss used in a finite-solver experiment is
$\mathcal L_K=\mathcal L(r_\psi(z_K),y)$, where $r_\psi$ is the task readout
and $y$ is the target. Reverse differentiation through the unrolled solver
obeys
\begin{align}
\bar z_K &= \nabla_{z_K}\mathcal L_K,
\label{eq:finite-backward-terminal}\\
\bar z_k
&= J_{\mathcal T_{\alpha,\theta}}(z_k,x)^\top\bar z_{k+1},
\qquad k=K-1,\ldots,0.
\label{eq:finite-backward-state}
\end{align}
Here $\bar z_k$ is the adjoint gradient with respect to the $k$th solver state,
and $J_{\mathcal T_{\alpha,\theta}}(z_k,x)$ is the Jacobian of the damped
update map with respect to $z$ at that iterate. The parameter and input
gradients accumulate over all solver steps,
\begin{align}
\nabla_\theta\mathcal L_K
&= \sum_{k=0}^{K-1}
\left(\partial_\theta\mathcal T_{\alpha,\theta}(z_k,x)\right)^\top
\bar z_{k+1},
\label{eq:finite-backward-params}\\
\nabla_x\mathcal L_K
&= \sum_{k=0}^{K-1}
\left(\partial_x\mathcal T_{\alpha,\theta}(z_k,x)\right)^\top
\bar z_{k+1}.
\label{eq:finite-backward-input}
\end{align}
In the evaluated SILVA models, parameters enter the solver through stimulus,
local-interaction, global-interaction, and normalization maps, while the task
loss also differentiates the final readout $r_\psi$. Since
$\partial_\theta\mathcal T_{\alpha,\theta}=\alpha\,\partial_\theta f_\theta$
for parameters entering the update map, Eqs.~\eqref{eq:finite-backward-state}
--\eqref{eq:finite-backward-input} are the exact chain rule for the finite
Picard loop used in those experiments.

\paragraph{Implicit adjoint backward.}
For the damped equilibrium condition
\(\mathcal T_\alpha(z,x)=(1-\alpha)z+\alpha f_\theta(z,x)=z\), the implicit
function theorem gives
$J_{\mathcal T} = (1-\alpha)I + \alpha J_{f_\theta}$. Substituting into
$I-J_{\mathcal T}$ term by term,
\begin{align}
I - J_{\mathcal T} &= I - \big[(1-\alpha)I + \alpha J_{f_\theta}\big],
\label{eq:IminusT-expanded}\\
I - J_{\mathcal T} &= \alpha\big(I - J_{f_\theta}\big),
\label{eq:IminusT}
\end{align}
Here $J_{\mathcal T}$ is the Jacobian of the damped solver map
$\mathcal T_\alpha$ with respect to $z$, $J_{f_\theta}$ is the corresponding
Jacobian of $f_\theta$, and $I$ is the identity on the flattened state space.
All Jacobians in this subsection are evaluated at the fixed point unless
otherwise stated.
Consequently,
$(I-J_{\mathcal T})^{-1} = \tfrac1\alpha(I-J_{f_\theta})^{-1}$ whenever
$I-J_{f_\theta}(z^\ast)$ is invertible, which Eq.~\eqref{eq:contraction}
guarantees. Applying the implicit function theorem to
\(\mathcal T_\alpha(z^\ast,x)=z^\ast\) with respect to any input or parameter
quantity $\xi$ entering through $f_\theta$, differentiating both sides gives
$(I-J_{\mathcal T})\,\partial z^\ast/\partial \xi =
\alpha\,\partial f_\theta/\partial \xi$, hence
\begin{equation}
\frac{\partial z^\ast}{\partial \xi} = \big(I-J_{f_\theta}\big)^{-1}\frac{\partial f_\theta}{\partial \xi},
\label{eq:alpha-cancel}
\end{equation}
Here $\xi$ denotes any input, weight, or other differentiable quantity entering
the update map through $f_\theta$. The derivatives are understood in vectorized
state coordinates, so $\partial z^\ast/\partial \xi$ is the sensitivity of the
fixed point to that quantity. The damping coefficient $\alpha$ cancels exactly.
The fixed point $z^\ast$ itself does not
depend on $\alpha$ (the fixed-point condition holds for any $\alpha\neq0$), so
its derivative cannot depend on $\alpha$ either. Damping changes only the
solver trajectory used to reach $z^\ast$, not the destination. This specializes
the general equilibrium-gradient result of \cite{bai2019deq}, stated there for
the undamped root condition $g_\theta(z)=f_\theta(z,x)-z=0$ directly. The
derivation above shows the same conclusion survives introducing the damped
iteration of Eq.~\eqref{eq:picard} as an explicit solver choice. The damping
coefficient is absent from the resulting gradient. The backward pass then
solves the $\alpha$-free adjoint system
\begin{equation}
\left(I - J_{f_\theta}(z^\ast)^\top\right)\Delta = \nabla_{z^\ast}\mathcal{L}
\label{eq:adjoint}
\end{equation}
Here $\Delta$ is the adjoint vector and $\nabla_{z^\ast}\mathcal L$ is the
gradient of the loss with respect to the fixed-point state. The corresponding
fixed-point gradients are
\begin{align}
\nabla_\theta\mathcal L
&= \left(\partial_\theta f_\theta(z^\ast,x)\right)^\top\Delta,
\label{eq:implicit-param-gradient}\\
\nabla_x\mathcal L
&= \left(\partial_x f_\theta(z^\ast,x)\right)^\top\Delta.
\label{eq:implicit-input-gradient}
\end{align}
Here $\partial_\theta f_\theta$ collects the derivatives of the stimulus,
local-interaction, global-interaction, and normalization maps with respect to
their parameters. Equivalently, one may solve
$(I-J_{\mathcal T_\alpha}(z^\ast)^\top)\mu=\nabla_{z^\ast}\mathcal L$ and
apply $(\partial_\theta\mathcal T_{\alpha,\theta})^\top\mu$. Because
$I-J_{\mathcal T_\alpha}=\alpha(I-J_{f_\theta})$ and
$\partial_\theta\mathcal T_{\alpha,\theta}=\alpha\partial_\theta f_\theta$,
the damping factors cancel in the exact fixed-point gradient. This is the
memory-efficient equilibrium-gradient identity used by DEQs
\cite{bai2019deq}. It requires the fixed point and Jacobian-vector products,
rather than the stored forward trajectory. The exact identity above is the
reference DEQ gradient. The adjoint-trained MNIST diagnostic uses the
GMRES adjoint described in the Supplementary Methods, with hatted gradients
used there to distinguish the executed adjoint path from the formal identity in
Eqs.~\eqref{eq:adjoint}--\eqref{eq:implicit-input-gradient}.
The remaining experiments use truncated backpropagation through the unrolled
$K$-step Picard loop, a weight-tied-network training regime that does not carry
the $O(1)$-memory guarantee.

\paragraph{Static global-matrix adjoint.} For a single
learned, non-gated global matrix $A$ (Section~\ref{sec:global}),
Eq.~\eqref{eq:adjoint} can be solved for $\nabla_A\mathcal L$ in closed
form. Using the receiver-first column convention of Eq.~\eqref{eq:propagator},
write the fixed point as a root condition on $b^\ast$ (pre-activation state,
$z^\ast=\tanh(b^\ast)$),
\begin{align}
g(b^\ast, A) &= b^\ast - m - A\tanh(b^\ast) = 0,
\label{eq:root}\\
m &= W_{\mathrm{stim}}x.
\label{eq:root-stimulus}
\end{align}
Here $b^\ast,m\in\mathbb R^N$, $A\in\mathbb R^{N\times N}$, and $\tanh$ is
applied elementwise. The root function $g(b^\ast,A)$ is zero exactly when the
pre-activation equilibrium condition is satisfied.
Contraction follows from $\tanh$'s 1-Lipschitz property,
$|\tanh(u)-\tanh(v)|\le|u-v|$. For any $b_1,b_2$,
\begin{align}
\|f_\theta(b_1,x)-f_\theta(b_2,x)\|
&= \|A(\tanh(b_1)-\tanh(b_2))\|,
\label{eq:tanh-lipschitz-exact}\\
\|f_\theta(b_1,x)-f_\theta(b_2,x)\|
&\le \|A\|_2\,\|b_1-b_2\|,
\label{eq:tanh-lipschitz}
\end{align}
Here $b_1$ and $b_2$ are two candidate pre-activation states, $\|\cdot\|$ is
the Euclidean vector norm, and $\|A\|_2$ is the spectral norm of the interaction
matrix.
Thus $\|A\|_2<1$ is a checkable-at-initialization sufficient condition for
this case to be a contraction, stronger than Eq.~\eqref{eq:contraction}'s
exact criterion (the operator norm dominates the spectral radius for
non-normal $A$, so this bound can be conservative. The same pattern appears in
Figure~\ref{fig:configspace} showing several configurations with $\rho$
comfortably below 1 at initialization despite no explicit norm constraint
on $A$). Totally differentiating $g=0$ at fixed $m$, with
$J=A\,\mathrm{diag}(1-\tanh^2(b^\ast))$,
\begin{align}
dg &= (I-J)\,db^\ast - dA\,\tanh(b^\ast) = 0,
\label{eq:total-diff}\\
db^\ast &= (I-J)^{-1}dA\,\tanh(b^\ast).
\label{eq:solved-total-diff}
\end{align}
Here $dg$, $db^\ast$, and $dA$ are differentials of the root function, fixed
point, and interaction matrix. The matrix
$J=A\,\mathrm{diag}(1-\tanh^2(b^\ast))$ is the Jacobian of
$A\tanh(b^\ast)$ with respect to $b^\ast$.
Given the loss gradient $v=\nabla_{b^\ast}\mathcal L$, the adjoint vector
$u$ solves $(I-J)^\top u=v$, the same linear solve as Eq.~\eqref{eq:adjoint}
specialized to this static matrix case. The parameter gradient follows by
substituting Eq.~\eqref{eq:total-diff} into
$d\mathcal L = v^\top db^\ast$.
\begin{equation}
\nabla_A\mathcal{L} = u\,\tanh(b^\ast)^\top.
\label{eq:adjoint-outer}
\end{equation}
The right-hand side is an outer product, so the resulting gradient has the same
$N\times N$ shape as $A$. Its $(i,j)$ entry is the sensitivity of the loss to
the influence from source node $j$ to receiver node $i$. The equivalent
row-oriented tensor convention writes the same interaction as
$\phi(b^\ast)A_{\mathrm{row}}$, which stores the transpose of the
receiver-first convention above. In that notation, the per-example gradient is
$\phi(b^\ast)^\top u$ before summing over a minibatch or differentiating
through an attention-generated $A_{\mathrm{row}}$.

When the global matrix is itself generated from the stimulus, the
outer-product adjoint first gives the sensitivity with respect to
$A_{\mathrm{global}}$. Differentiating the attention map
$A_{\mathrm{global}}(m)$ then propagates this sensitivity to the stimulus and
to the parameters that define the global interaction. The resulting adjoint is
the sum of the direct stimulus contribution and the contribution mediated by
the global-attention map.

\paragraph{Head-averaged attention adjoint.}
This case applies to local-attention layers in which the $H$ head outputs are
averaged to a single per-node value, including the chemistry configuration with
averaged rather than concatenated heads and the adjoint-differentiated layer. Let
$\mathcal A_\varphi(y)=\{\eta_{i,h}\}_{i,h}$ denote the multi-head attention map
before head aggregation, where $\varphi$ are the attention parameters and
$\eta_{i,h}$ is the output of head $h$ at node $i$. The averaged local output is
\begin{align}
\ell_i &= \frac{1}{H}\sum_{h=1}^{H}\eta_{i,h},
\label{eq:head-average}\\
\bar\eta_{i,h} &= \frac{1}{H}\bar\ell_i,
\qquad i=1,\ldots,N,\quad h=1,\ldots,H,
\label{eq:head-average-adjoint}\\
\bar y_{\mathrm{local}}
&= \left(D_y\mathcal A_\varphi(y)\right)^\top \bar\eta,
\label{eq:head-average-input-adjoint}\\
\nabla_\varphi\mathcal L
&= \left(D_\varphi\mathcal A_\varphi(y)\right)^\top \bar\eta.
\label{eq:head-average-param-adjoint}
\end{align}
Here $\bar\ell_i=\partial\mathcal L/\partial\ell_i$ is the incoming adjoint for
the averaged local output, $\bar\eta_{i,h}=\partial\mathcal L/\partial
\eta_{i,h}$ is the adjoint for an individual head output, and
$\bar y_{\mathrm{local}}$ is the local-attention contribution to the adjoint of
the signal $y$. The matrices $D_y\mathcal A_\varphi$ and
$D_\varphi\mathcal A_\varphi$ are the Jacobians of the multi-head attention map
with respect to its input signal and parameters.
Thus head averaging contributes only the factor $1/H$ in
Eq.~\eqref{eq:head-average-adjoint}; the remaining local adjoint is the standard
vector--Jacobian product of the attention map. The local and global
contributions then combine as in Eq.~\eqref{eq:adjoint}. The global term is
unaffected by the head-averaging reduction. Citation-network and CLUSTER layers
use concatenated attention heads in the finite unrolled training path, so their
gradients are governed by
Eqs.~\eqref{eq:finite-backward-state}--\eqref{eq:finite-backward-params}.

\subsection{Spectral normalization}

Spectral normalization $W_{\mathrm{SN}} = W/\sigma_{\max}(W)$
\cite{miyato2018spectral} on a weight matrix appearing linearly in
$f_\theta$ makes that linear map non-expansive
($\|W_{\mathrm{SN}}z\|\le\|z\|$, by submultiplicativity of the operator
norm), so that each component force is individually bounded. Non-expansiveness
of individual linear maps does not by itself make their sum strictly
contractive. This mirrors the weight-norm-constrained view of stability used
to design provably stable deep and ordinary-differential-equation-inspired
architectures \cite{haber2017stable}, and the same argument as
Eq.~\eqref{eq:tanh-lipschitz} applied per weight matrix instead of to a single
static matrix. Here $\sigma_{\max}(W)$ is the largest singular value of $W$,
and $W_{\mathrm{SN}}$ is the normalized version used in the forward map. In
these experiments, spectral normalization is applied to the
molecular-regression model's global and stimulus weight matrices. Only the
state-dependent parts of the update contribute to the Jacobian with respect to
$z$, so the resulting stability is monitored through the damped-step diagnostic
$\rho_{\mathcal T}$ (Section~\ref{sec:rho}) rather than assumed from weight
normalization alone.

\section{Results}\label{sec:experiments}

\paragraph*{MNIST solver dynamics.}

\begin{figure}[h]
\centering
\includegraphics[width=1\textwidth]{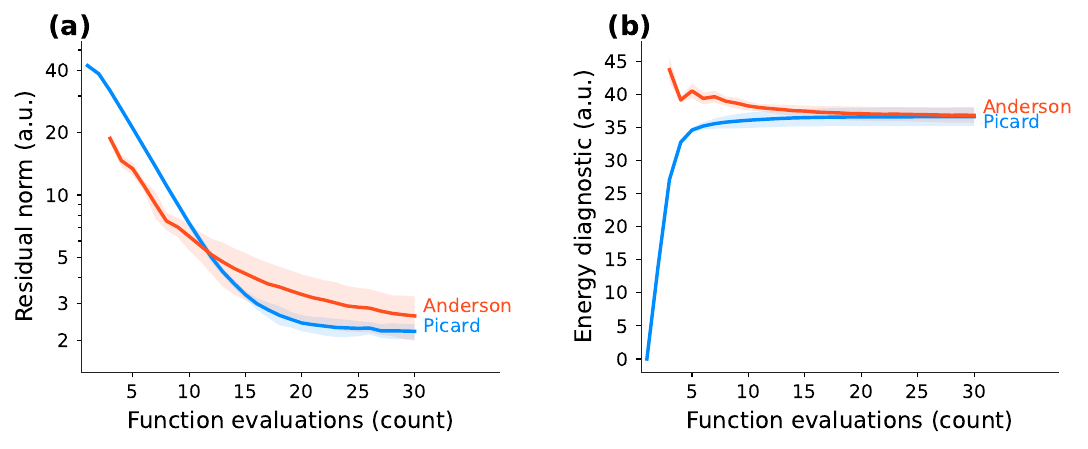}
\caption{Convergence diagnostics for trained SILVA image classifiers. (a)
Damped fixed-point residual $\|\mathcal T_\theta(z_k;x)-z_k\|_2$ over solver
function evaluations. (b) Energy diagnostic over the same iterations. Lines
show means and bands show standard deviations across five independently
trained MNIST models. Axes report function-evaluation counts and arbitrary
units for the residual and energy-diagnostic magnitudes.}
\label{fig:solverdynreal}
\end{figure}

Two MNIST training regimes are used for different purposes. The principal
finite-unrolled accuracy experiment uses a 90/10 train/validation split with a
held-out test set never used for model selection. A five-seed sweep of the
single-layer MNIST local+global architecture \cite{lecun1998gradient}
($\alpha=0.25$, $K=20$) reaches $97.23\pm0.13\%$ test accuracy.
Figure~\ref{fig:solverdynreal} compares Picard and Anderson solvers on the same
five trained models. In Figure~\ref{fig:solverdynreal}a, the Anderson trajectory
has already reduced the residual by its first plotted accelerated step and stays
below Picard through the early function evaluations. By 20--30 evaluations, both
methods reach the same low-residual regime, with Picard marginally lower at the
end of this sweep. This supports the interpretation that Anderson changes the
route to the fixed point rather than the learned equilibrium being evaluated.
Figure~\ref{fig:solverdynreal}b gives the paired solver-state diagnostic. Picard
rises from the initialized state, whereas Anderson relaxes from a higher initial
diagnostic value, and both trajectories approach the same plateau. The diagnostic
therefore tracks convergence toward a common trained state. It is not used as a
proof of global Lyapunov descent.

\phantomsection\label{sec:adjoint-mnist}
\paragraph*{Implicit-adjoint MNIST equilibrium.}

\begin{figure}[h]
\centering
\includegraphics[width=1\textwidth]{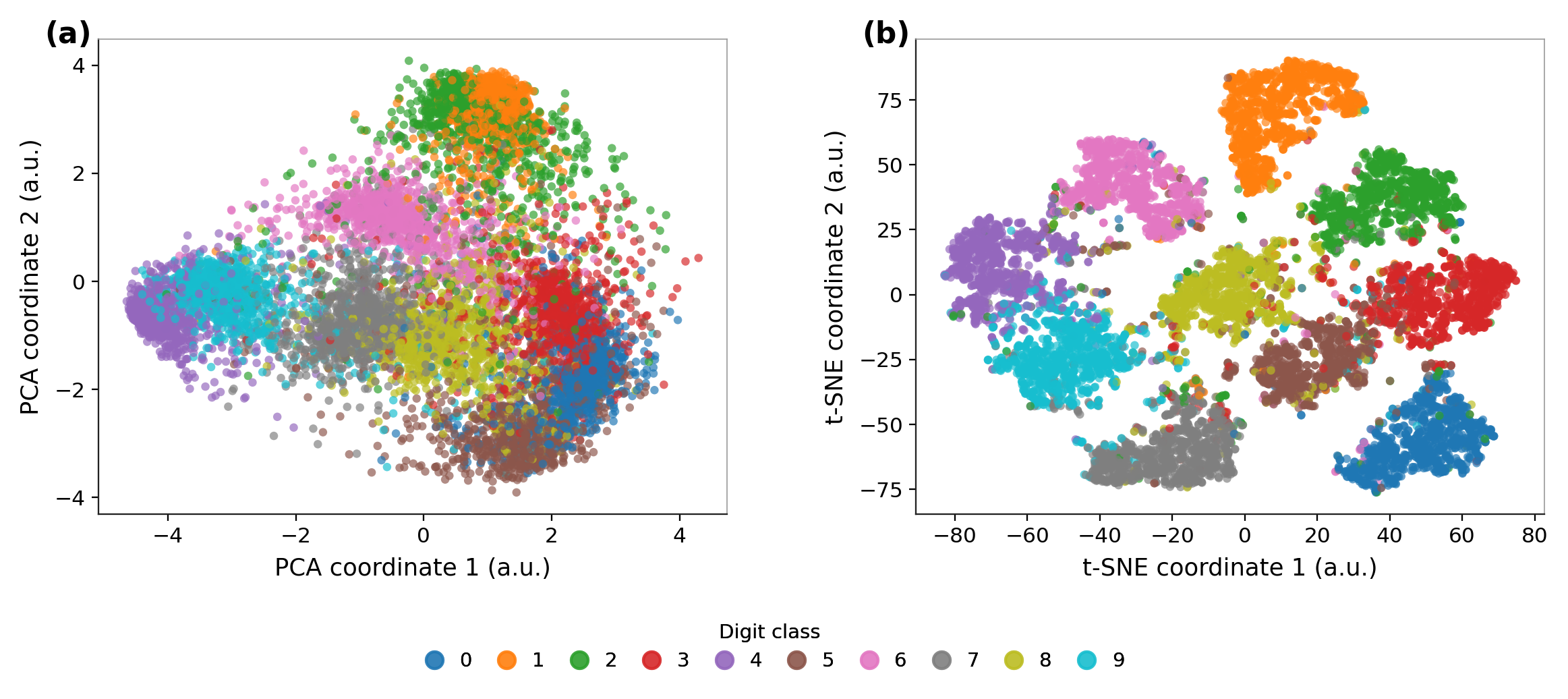}
\caption{Input-conditioned equilibrium-state geometry from the converged
hidden state $z^\ast$ on MNIST test data from the adjoint-trained diagnostic
model. (a) PCA projection. (b) t-SNE projection. Each point is one test image
represented by its terminal hidden state before readout. Points are colored by
digit class, and projection coordinates are reported in arbitrary units.}
\label{fig:attractorreal}
\end{figure}

A separate MNIST experiment trained with a GMRES adjoint, rather than
truncated backpropagation through the unrolled solver, learned a non-trivial
classifier but remained below the finite-unrolled MNIST configuration. It is
used here as an equilibrium-state separability diagnostic.
Figure~\ref{fig:attractorreal} shows principal component analysis (PCA) and
t-SNE projections \cite{jolliffe2002pca,vandermaaten2008tsne} of the
converged hidden state across the MNIST test set at 89.9\% validation
accuracy. For this diagnostic,
the fixed-point solver is run on each test image and the terminal state
$z^\ast$ is stored before the classification readout. PCA displays the two
largest variance directions of these full hidden states, while t-SNE displays
their local neighborhood organization in a nonlinear embedding. The plotted
points are therefore projections of vector attractors rather than logits or
input pixels. In
Figure~\ref{fig:attractorreal}a, the first two linear principal components
separate several digit regions but also retain substantial overlap. The
equilibrium state therefore contains class structure without collapsing into
ten linearly disjoint regions in this projection. In
Figure~\ref{fig:attractorreal}b, the nonlinear neighborhood embedding exposes
more coherent class neighborhoods, with residual mixing between visually related
or ambiguous classes. The silhouette score $0.382$ and Davies--Bouldin score
$1.098$ \cite{rousseeuw1987silhouettes,davies1979cluster} are computed on the
full $z^\ast$ vectors before two-dimensional projection. They quantify this
moderate, rather than complete, attractor separation.

\paragraph*{Stacked equilibrium dynamics on CIFAR-10.}

\begin{figure}[h]
\centering
\includegraphics[width=1\textwidth]{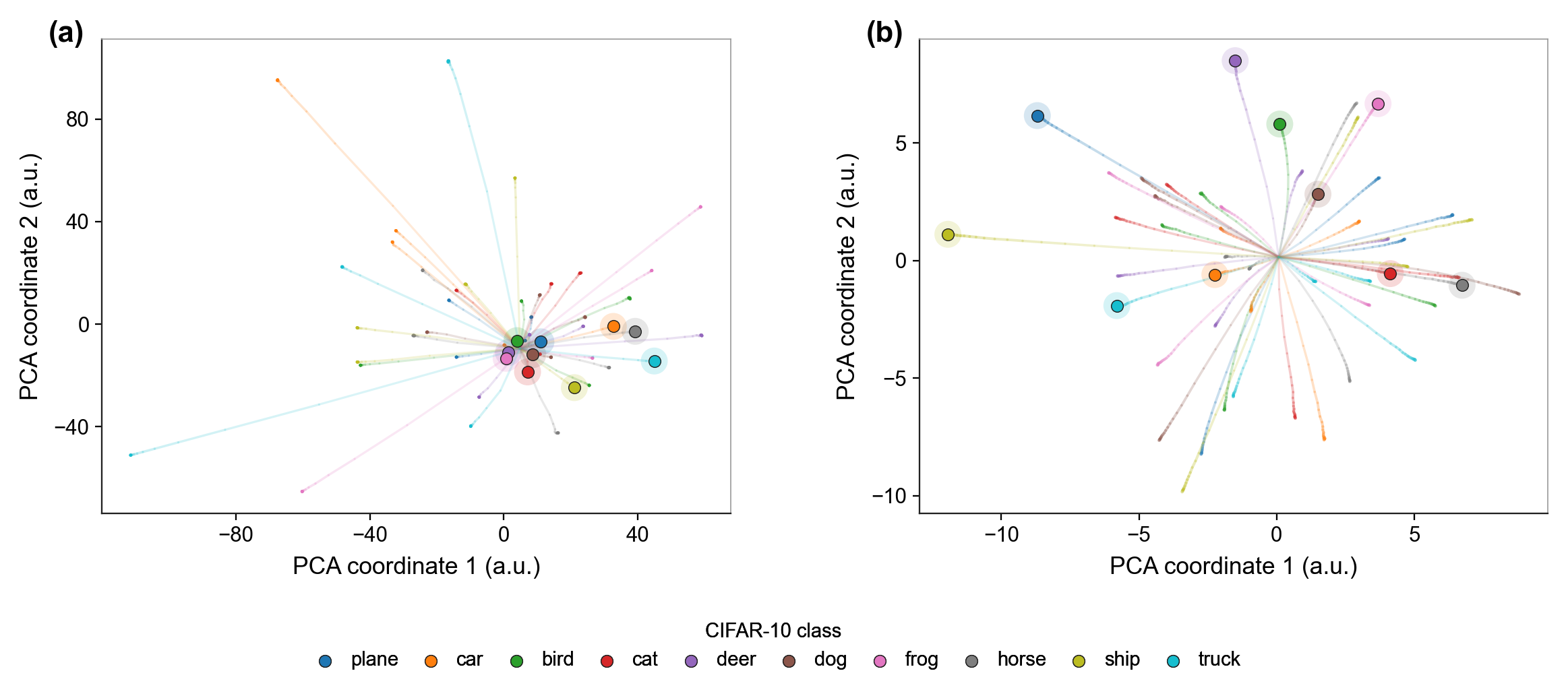}
\caption{Stacked equilibrium hierarchy on CIFAR-10. PCA-projected
Picard-iteration trajectories are shown for five independently-trained seeds
and one held-out image per class. Curves show solver-time hidden-state
trajectories, and large markers highlight representative terminal equilibria.
(a) Fast first layer with $\alpha=0.5$. (b) Slow second layer with
$\alpha=0.2$. PCA coordinates are reported in arbitrary units.}
\label{fig:hierarchyreal}
\end{figure}

The same damped fixed-point template, with the CIFAR-specific convolutional
input encoder (Section~\ref{sec:method}), reaches $74.77\pm1.59\%$ test
accuracy on CIFAR-10 \cite{krizhevsky2009cifar} over five independent seeds
at hidden width 64. Figure~\ref{fig:hierarchyreal} traces the
Picard-iteration trajectory, not only the converged endpoint, for both equilibrium
layers across all five trained seeds, one held-out test image per class. Each
curve follows the hidden state sequence $z_0,z_1,\ldots,z_K$ through solver
time after projection into the panel's PCA plane. Color indicates CIFAR-10
class. The large terminal markers highlight representative approximate
equilibria, and the length of a curve reports change in the projected hidden
state rather than distance in the image space. In
Figure~\ref{fig:hierarchyreal}a, the first layer with $\alpha=0.5$ moves through
a broad state-space range before settling, and the endpoints remain
class-dependent across seeds. In Figure~\ref{fig:hierarchyreal}b, the second
layer with $\alpha=0.2$ follows shorter trajectories in a more compact projected
region while preserving class-conditioned endpoints. The stacked model therefore
does not merely add depth. It applies a second equilibrium transform with a
different relaxation scale to the representation produced by the first.

\paragraph*{Molecular property regression on ZINC.}

Property-regression targets are drawn from the ZINC molecular-property benchmark
\cite{irwin2012zinc} (12k subset). Table~\ref{tab:zinc_baselines} gives
mean absolute error (MAE) for graph convolutional network (GCN), graph attention
network (GAT), graph isomorphism network (GIN), principal neighbourhood
aggregation (PNA), and gated graph convolutional network (GatedGCN) baselines
\cite{kipf2017gcn,velickovic2018gat,xu2019gin,corso2020pna,bresson2017residualgated,dwivedi2020benchmarking},
each trained under the same ZINC split, hidden width, and evaluation
protocol. The SILVA molecular-regression model
reaches $0.3893\pm0.0132$ validation MAE across five seeds
(Table~\ref{tab:zinc_baselines}), improving on GCN, GAT, GIN, and GatedGCN
within the same training and evaluation protocol while trailing PNA
($0.3475\pm0.0130$).
The SILVA model uses bond-type input through the molecular edge encoder,
whereas the baselines in this matched protocol use only the shared node-feature
and graph-topology setting. The ZINC table is therefore a same-split and
same-budget comparison. It does not isolate parameter count or bond-type edge
features from the SILVA interaction field.

\begin{table}[h]
\centering
\caption{ZINC-12k validation mean absolute error (MAE). Lower is better. Values are mean $\pm$ standard deviation across independent seeds. GCN denotes a graph convolutional network, GAT a graph attention network, GIN a graph isomorphism network, PNA principal neighbourhood aggregation, and GatedGCN a gated graph convolutional network. All rows share the same ZINC split, training protocol, and hidden dimension $d=64$. The SILVA row additionally uses bond-type edge input, and the rows are not parameter-matched.}
\label{tab:zinc_baselines}
\begin{tabular}{lccc}
\toprule
Model & MAE & $n$ seeds \\
\midrule
SILVA bond-aware & $0.3893 \pm 0.0132$ & 5 \\
GCN~\cite{kipf2017gcn} & $0.5509 \pm 0.0032$ & 5 \\
GAT~\cite{velickovic2018gat} & $0.5068 \pm 0.0093$ & 5 \\
GIN~\cite{xu2019gin} & $0.4744 \pm 0.0063$ & 5 \\
PNA~\cite{corso2020pna} & $0.3475 \pm 0.0130$ & 5 \\
GatedGCN~\cite{dwivedi2020benchmarking} & $0.4096 \pm 0.0067$ & 5 \\
\bottomrule
\end{tabular}
\end{table}

\phantomsection\label{sec:nodeclass-scope}
\paragraph*{Citation-network node classification.}

SILVA casts images, molecules, and citation graphs as interaction systems
solved to a fixed point under one structured template
(Section~\ref{sec:grand}). This representation predicts that the global
interaction term should be useful when a task needs information mixing beyond
immediate local structure and neutral when local structure already contains the
relevant signal. Node classification on citation networks tests this regime
directly, since these tasks are transductive,
single-graph, and trained on a very small labeled budget (140/120/60 nodes
for Cora/Citeseer/Pubmed respectively
\cite{sen2008collective,yang2016revisiting}). The standard
citation-network graph convolutional network (GCN) and graph attention network
(GAT) baselines were introduced in this same
semi-supervised, citation-edge aggregation setting
\cite{kipf2017gcn,velickovic2018gat}. These datasets therefore probe a
local-structure-dominated regime for the additional global mechanism.

Table~\ref{tab:node_classification} compares
graph convolutional network (GCN) and graph attention network (GAT) baselines
(no equilibrium solve, no
global term, five seeds each) against SILVA (local term
instantiated as graph attention over the real citation graph, global term
as a scalar-gated mean-field broadcast). On Cora, SILVA remains below both
local graph baselines and shows larger seed-to-seed
dispersion. SILVA reaches $70.80\pm4.86\%$ against $77.04\pm1.21\%$
(graph convolution) and $78.48\pm0.95\%$ (graph attention). Citeseer shows the
same ordering, with SILVA at
$55.96\pm1.98\%$ against $67.32\pm1.22\%$ and $67.60\pm1.36\%$. Together,
Cora and Citeseer indicate that, in small local-structure-dominated
transductive regimes, the additional equilibrium and global machinery does not
improve over a plain graph-attention classifier with no fixed-point solve. The
corresponding longer-range test is CLUSTER \cite{dwivedi2020benchmarking}.
Published 4-to-16-layer graph-convolution
baselines without positional encoding span approximately 47.8--69.0\%, whereas
gated graph convolution spans approximately 60.4--73.8\% under the same
protocol, as discussed in the CLUSTER result below.

A direct channel-attention global term, obtained by applying the vision
attention structure to the node dimension of a full graph, collapses at graph
scale to an exactly uniform attention distribution (measured row-entropy ratio
1.0000) and accuracy near chance. The scalar-gated mean-field broadcast of
Section~\ref{sec:global} removes this collapse by construction. A bounded
top-$k$ attention variant uses genuine node-to-node attention restricted to
each node's $k$ nearest candidates. Across five seeds it reaches
$69.12\pm4.95\%$ on Cora and $55.80\pm3.53\%$ on Citeseer, within the same
range as the mean-field variant.

\begin{table}[h]
\centering
\caption{Citation-network node-classification validation accuracy. Higher is better. Values are mean $\pm$ standard deviation across $n=5$ independent seeds on the public train/validation/test splits \cite{yang2016revisiting} matching Table~1 of \cite{velickovic2018gat}. GCN denotes graph convolutional network and GAT denotes graph attention network. Pubmed SILVA is shown per seed, reflecting bimodal outcomes.}
\label{tab:node_classification}
\begin{tabular}{lccc}
\toprule
Dataset & GCN~\cite{kipf2017gcn} & GAT~\cite{velickovic2018gat} & SILVA \\
\midrule
Cora & $0.7704 \pm 0.0121$ & $0.7848 \pm 0.0095$ & $0.7080 \pm 0.0486$ \\
Citeseer & $0.6732 \pm 0.0122$ & $0.6760 \pm 0.0136$ & $0.5596 \pm 0.0198$ \\
Pubmed & $0.7924 \pm 0.0107$ & $0.7852 \pm 0.0097$ & \begin{tabular}[c]{@{}c@{}}$0.706, 0.746, 0.770$\\$0.416, 0.388$\end{tabular} \\
\bottomrule
\end{tabular}
\par\smallskip\noindent\emph{Note. Pubmed SILVA lists three well-converged seeds on the first line and two collapsed seeds on the second.}
\end{table}

\begin{figure}[h]
\centering
\includegraphics[width=1\textwidth]{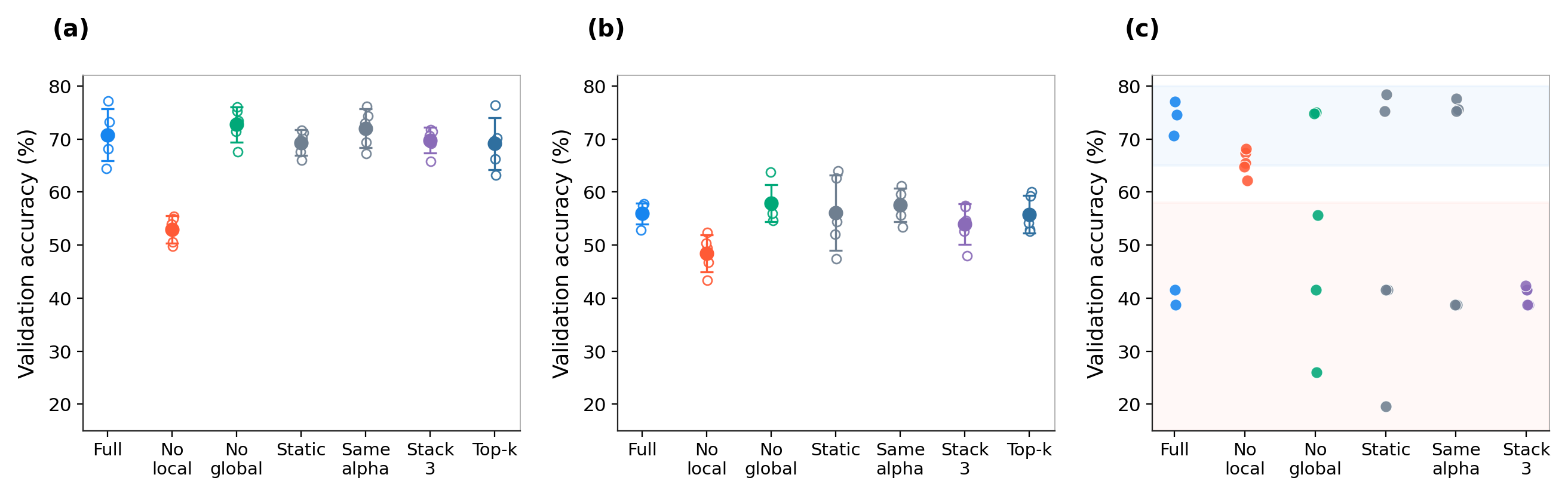}
\caption{Citation-network ablations. (a) Cora. (b) Citeseer. Filled markers
and whiskers show mean $\pm$ standard deviation across five seeds, with
individual seeds overlaid as open markers. (c) Individual Pubmed seed values
for evaluated variants.}
\label{fig:silvaablation}
\end{figure}

\textbf{Term-level attribution.} A seven-variant ablation
study on Cora and Citeseer (Figure~\ref{fig:silvaablation}, five seeds each) sharpens this
finding. Figure~\ref{fig:silvaablation}a shows that on Cora, removing the local
term is the only ablation that causes a large accuracy drop
($70.8\%\to53.0\%$), whereas removing the global term slightly exceeds the full
model ($72.7\%$ vs.\ $70.8\%$). Figure~\ref{fig:silvaablation}b repeats the
same pattern on Citeseer. Removing the local term again collapses accuracy
($56.0\%\to48.5\%$), whereas the no-global model again slightly exceeds the full
model ($57.9\%$ vs.\ $56.0\%$). The static global matrix, equal fast/slow
timescales, three-layer stack, and top-$k$ mechanism all remain near the full
model on these two datasets. The two panels therefore attribute the
citation-network limitation to the global term not adding useful information,
not to the absence of local graph processing.

On Pubmed, the evaluated variants do not form a single summary pattern and
are therefore shown per seed. Figure~\ref{fig:silvaablation}c separates the
well-converged seeds from collapsed seeds visually rather than compressing them
into a single mean. Four variants, the full model, no-global, static, and
same-alpha, split across their five seeds into a well-converged band
(70--78\% accuracy) and a collapsed band (19--56\% accuracy). The remaining
evaluated variants show different behavior. The three-layer stack collapses
uniformly, all five seeds landing in the 39--42\% range with none reaching the
well-converged band. The no-local (global-only) variant instead converges
tightly and unimodally in an intermediate 62--68\% range and falls in neither
band. The Pubmed panel therefore reveals a seed-dependent convergence
bifurcation, not ordinary statistical noise around one operating point. The
bimodal cases indicate a convergence instability specific to Pubmed's
substantially larger graph ($N=19{,}717$ vs.\ Cora's $N=2{,}708$), plausibly
connected to the spectral-radius framing of Section~\ref{sec:pathsum}. The
uniformly-collapsed and uniformly-converged exceptions indicate this instability
is not a fixed property of the Pubmed graph alone, but depends on which
interaction terms are active.

\phantomsection\label{sec:cluster}
\paragraph*{Long-range node classification on CLUSTER.}

On CLUSTER \cite{dwivedi2020benchmarking}, the controlled comparison is between two
baselines trained under the same data split and evaluation protocol, but not
matched parameter counts. A graph convolutional network (GCN) reaches
$53.22\pm0.50\%$ across five seeds, and a gated graph convolutional network
(GatedGCN) reaches
$58.75\pm0.23\%$ across five seeds. These baselines establish a consistent
comparison gap in the direction the benchmark is designed to produce. Dwivedi et
al.~\cite{dwivedi2020benchmarking} report two distinct graph-convolution baselines under this same protocol,
differing in normalization convention. They report a $47.8$--$69.0\%$ range (4-to-16
layers, their ``GCN'' row) and a $53.4$--$68.5\%$ range (their ``vanilla
GCN'' row). The graph-convolution baseline sits almost exactly on the
latter's 4-layer point ($53.22\%$ vs.\ $53.45\%$), indicating that the
configuration follows that convention. For gated graph convolution,
their plain (no positional encoding) range is $60.4$--$73.8\%$. A
positional-encoding-augmented variant reaches $76.1\%$
at 16 layers but is architecturally distinct from the plain 4-layer
gated-convolution baseline used here. Across those published protocol and
hyperparameter settings, the
graph-convolution baseline sits
within its matching published range, and the gated graph-convolution
result sits just below its published range while preserving the same
qualitative ordering. Because CLUSTER is designed to reward a longer-reach
mechanism, it directly tests whether the full local+global mechanism can help
under longer-range demands, complementing the local-structure-dominated regime
identified in the citation-network analysis above. The SILVA local+global
mechanism, trained under the identical 50-epoch budget as the two baselines
above, reaches $73.04\pm0.60\%$ validation accuracy across five independent
seeds. It exceeds the same-budget graph-convolution rerun by $19.82$
percentage points and the same-budget gated graph-convolution rerun by $14.29$
percentage points. A
four-seed ablation that removes the global interaction term while
keeping the local interaction term reaches $67.55\pm0.84\%$, isolating a
$5.49$ percentage-point aggregate difference associated with the global term.
The CLUSTER benchmark is the evaluated setting in which the global
interaction term is load-bearing.

\begin{table}[h]
\centering
\caption{CLUSTER node-classification validation accuracy over synthetic graph
instances. Values are mean
$\pm$ standard deviation across independent seeds. GCN denotes graph
convolutional network and GatedGCN denotes gated graph convolutional network.}
\label{tab:clusterresults}
\begin{tabular}{lcc}
\toprule
Model & Validation accuracy & $n$ seeds \\
\midrule
GCN & $53.22\pm0.50\%$ & 5 \\
GatedGCN & $58.75\pm0.23\%$ & 5 \\
SILVA no global & $67.55\pm0.84\%$ & 4 \\
SILVA full & $73.04\pm0.60\%$ & 5 \\
\bottomrule
\end{tabular}
\end{table}

\phantomsection\label{sec:stack3}
\paragraph*{Stacked equilibrium hierarchy.}

The two-layer fast/slow stack (Eq.~\eqref{eq:picard}) used in the CIFAR-10,
ZINC, citation-network, and CLUSTER experiments is the shallowest
instance of a more general architectural idea (Section~\ref{sec:method}). It is
a hierarchy of equilibria at progressively slower timescales, each still solving
the full interaction structure. This composition differs from previously
evaluated recurrent hierarchies by making every rung an implicit local+global
equilibrium. It is a
multi-timescale hierarchy \cite{koutnik2014clockwork,elhihi1995hierarchical}
in which every rung is an implicit fixed-point solve with its own local+global
interaction structure, not a single fixed-depth recurrent step.

A first three-layer instance ($\alpha=0.5,0.35,0.2$) was evaluated over five
independent seeds on Cora and Citeseer under the same data split and training
budget as Table~\ref{tab:node_classification}. It reaches
$69.76\pm2.43\%$ on Cora, within the two-layer variant's
$70.80\pm4.86\%$ spread, and $53.96\pm3.88\%$ on Citeseer, below the
two-layer variant's $55.96\pm1.98\%$. The seed-0 traces show training loss
continuing to fall while validation loss climbs past approximately epoch 30.
This overfitting signature is consistent with the citation-network analysis
above, which identifies citation networks as a regime in which additional
equilibrium capacity is not expected to help.

Depth can also be added \emph{within} a single equilibrium layer's own solver
iteration by repeating the local term's message-passing step
$\mathrm{local\_depth}$ times, weight-tied, before combining with the global
term. This differs from adding separate stacked equilibria as above. A Cora
sweep with one seed per local-depth-2 variant (six variants at
$\mathrm{local\_depth}=2$, plus the full model at
$\mathrm{local\_depth}=3$) found accuracy decreasing at every variant
tested. The deltas compare the five-seed means at
$\mathrm{local\_depth}=1$ with the single completed seed at
$\mathrm{local\_depth}=2$. The largest decreases occur for the top-$k$ global-term
variant ($69.1\%\to55.4\%$) and the combination with three-layer stacking
($69.8\%\to63.4\%$), consistent with the over-smoothing failure mode identified
for deeper graph convolutional propagation \cite{li2018deeper}.
Added local depth is therefore unfavorable in this citation-network setting.

\phantomsection\label{sec:multidomain}
\paragraph*{Comparison with multi-domain latent-attention architectures.}

Several architectures target multiple, heterogeneous data domains with one
mechanism. Perceiver IO \cite{jaegle2022perceiverio}
is a single cross-attention architecture demonstrated across images
(ImageNet, 84.5\% top-1), language, optical flow, multimodal audio-video
tasks, and symbolic game state, by mapping arbitrary inputs into a
fixed-size latent array via cross-attention and querying that latent
flexibly for arbitrary outputs. Graph Perceiver IO
\cite{bae2022graphperceiverio} extends the same architecture to
graph-structured data and to joint image-graph few-shot classification.

Perceiver IO shares no benchmark with this evaluation suite, so the
comparison is architectural. Graph Perceiver IO overlaps with SILVA on the same
three citation-network datasets evaluated in
the citation-network analysis above under the public split and 100-repeat protocol.
It reports Cora $83.9\pm0.6\%$, Citeseer $70.1\pm1.0\%$, Pubmed $79.9\pm0.4\%$
\cite{bae2022graphperceiverio}, exceeding both the graph-convolution/graph-attention
baselines and the SILVA values in Table~\ref{tab:node_classification}. Its reported baselines on these
splits are drawn from literature-tuned values rather than a matched rerun under
the present protocol. The accuracy gap therefore reflects both architectural
and protocol differences.

On citation networks, Graph Perceiver IO is ahead under its reported protocol.
The citation-network analysis above identifies citation networks as a
local-structure-dominated regime for which the additional global machinery is
weakly expressed. SILVA contributes a falsifiable dynamical account of the
mechanism itself. This account includes a monitored
spectral radius that tracks instability (Section~\ref{sec:rho}), a
solver-state energy diagnostic and local descent surrogate
(Section~\ref{sec:energy}), a finite path-sum
decomposition for the local linearization (Section~\ref{sec:pathsum}),
and an explicit, graph-structured local/global interaction split in place of
an opaque learned latent bottleneck.

\phantomsection\label{sec:ablations}
\paragraph*{Local--global interaction ablations.}

\begin{figure}[h]
\centering
\includegraphics[width=1\textwidth]{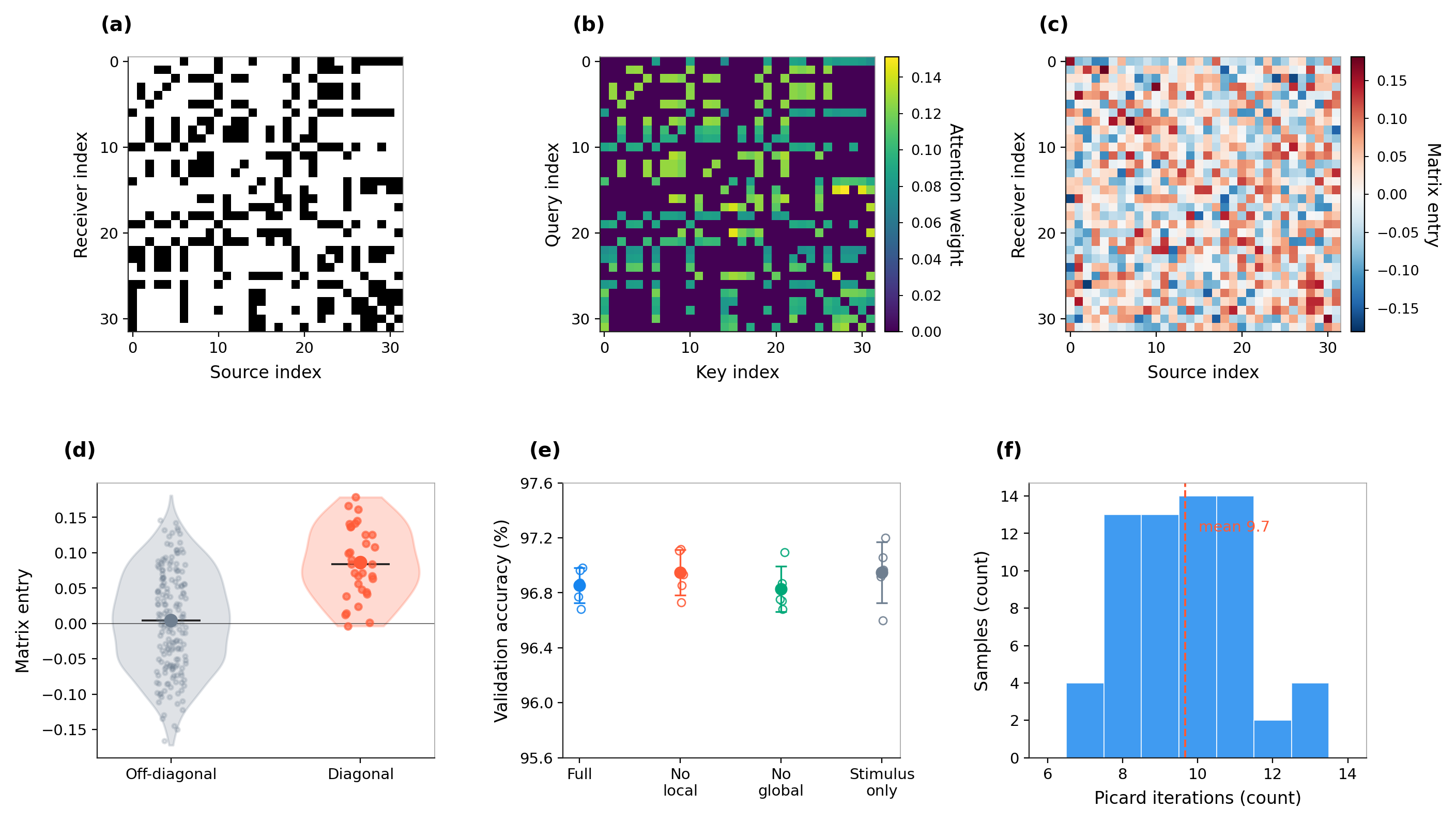}
\caption{MNIST local-global diagnostics. (a--d,f) Operator and solver
diagnostics from an adjoint-trained equilibrium model with local attention and
a learned static global matrix.
(a) Learned $k$-nearest-neighbor graph in hidden-neuron space for one test
sample. (b) Head-averaged graph-attention weights for the same sample and
layer. (c) Learned static global interaction matrix. (d) Distributions of
diagonal and off-diagonal entries of the static global matrix, with filled
markers denoting means. (e) Four-arm MNIST ablation across full, local-only,
global-only, and stimulus-only variants, shown as mean $\pm$ standard
deviation with individual seeds overlaid. (f) Picard iterations required to
reach residual tolerance $10^{-3}$ over 64 test samples.}
\label{fig:interactionstructure}
\end{figure}

A four-arm ablation on MNIST (full local+global model, local only, global
only, and a stimulus-only variant with neither interaction term, five seeds each,
matched hidden width and training
budget) reaches $96.85\pm0.13\%$, $96.83\pm0.16\%$, $96.95\pm0.17\%$, and
$96.95\pm0.22\%$ validation accuracy respectively. The four arms overlap within
seed variation, with every pairwise difference within 1--2 standard deviations.
A direct diagnostic confirms that the local and global terms alter the
computation when enabled. The maximum output difference between the full and
stimulus-only variants is approximately 1.01 at identical initialization and
input. The near-identical accuracy therefore reflects this task and parameter
budget rather than absence of a computational difference between variants.
Figure~\ref{fig:interactionstructure}e shows the full comparison. Together with
the citation-network result above, this extends the same finding to a second
domain.
At this capacity budget, MNIST classification is easy enough that the
additional local/global interaction machinery does not measurably improve
on the stimulus-only equilibrium variant, consistent with the representation's own
premise that these terms are expected to help specifically when a task
needs richer structure than a single injected stimulus already provides.
The remaining panels of Figure~\ref{fig:interactionstructure} provide an
operator-level MNIST diagnostic from the adjoint-trained equilibrium model with
local attention and a learned static global matrix.
Figure~\ref{fig:interactionstructure}a shows the learned hidden-neuron
$k$-nearest-neighbor graph used by the local term. Its
sparse binary structure fixes which state coordinates may exchange local
messages. Figure~\ref{fig:interactionstructure}b shows the corresponding
head-averaged attention weights for one sample. The weights are nonuniform over
the allowed edges, showing that the local term is an input-dependent interaction
operator rather than a fixed averaging rule. Figure~\ref{fig:interactionstructure}c
shows the static global matrix. Its dense signed entries complement the sparse
local graph by allowing every coordinate to contribute to every other coordinate
through the global channel. Figure~\ref{fig:interactionstructure}d compares the
diagonal and off-diagonal entry distributions of this matrix. The larger
positive diagonal mean indicates an emergent self-feedback component inside the
learned global interaction. Figure~\ref{fig:interactionstructure}f closes the
diagnostic by showing that the Picard solver reaches tolerance in a small
finite number of iterations on the tested samples. Together, these panels show
the concrete objects represented by the SILVA decomposition: sparse local
support, learned local weights, dense global coupling, self-interaction in the
global operator, and finite solver effort.

\textbf{Parameter-matched computational cost.} The ablation above holds
the equilibrium solve fixed and varies only which interaction terms feed it.
A single-hidden-layer feedforward network with no fixed-point
iteration, no local or global interaction term, and a parameter count
matched to the full equilibrium model to within 0.4\% (63,610 vs.\ 63,370
parameters) reaches $97.29\pm0.23\%$ validation accuracy across five
seeds, matching or marginally exceeding the full model's
$96.85\pm0.13\%$, at 2.5- to 6.5-fold lower wall-clock cost per epoch
(12--29s vs.\ 72--80s, measured under identical conditions). Taken
together with the four-arm ablation above, this indicates that at the
MNIST task and parameter budget tested here, neither the choice of
interaction term nor the equilibrium mechanism itself is measurably
earning its additional computational cost over the feedforward
baseline. The result identifies a task regime in which the SILVA mechanism is
unnecessary at the tested capacity. Together with the citation-network result
above, it supports the interpretation that the representation's value lies in
tasks that need richer structure than a single-shot stimulus already provides,
and that MNIST classification at this capacity is not such a task.

\textbf{Linear decodability of frozen representations.} Classification accuracy requires the
minimal statistic sufficient to separate ten classes, and has no reason to
reward any additional structure a representation happens to preserve. Two
linear probes test this on the trained frozen representations, with no
reference to digit identity. The probes are a
four-way rotation probe (0/90/180/270 degrees, exact array rotation, no
interpolation) and a strictly linear pixel-space reconstruction decoder,
both trained only on top of the frozen representation with the backbone
untouched. On both probes the plain feedforward baseline's representation
is more linearly decodable than the full equilibrium model's. Rotation
probe accuracy $80.3\%$ vs.\ $74.6\%$ (chance $25\%$), and reconstruction
mean squared error $0.0284$ vs.\ $0.0318$ (lower is better). The
comparison differs in representation width (80 vs.\
64 dimensions, matched on parameter count rather than dimensionality,
since the feedforward network concentrates its budget into one wide layer
where the equilibrium model spreads it across several weight matrices),
so some of this gap may reflect probe capacity rather than information
content. These probes do not support the hypothesis that
the interaction terms incidentally preserve more usable structure than a
plain baseline, on the two probes tested here.

\textbf{Optimization diagnostics.} Mean validation accuracy across seeds
can hide whether a flat ablation reflects task structure or optimization
failure. An optimization explanation would predict a landscape with
variable-quality local optima, the
same signature already observed for the SILVA
node-classification variants on Pubmed in the citation-network analysis above,
where seeds split bimodally between well-converged and collapsed outcomes.
Two checks distinguish these on the MNIST ablation matrix specifically.
First, final training losses across the four arms overlap tightly and remain
close to zero (final training loss
$0.0147$--$0.0205$, no arm systematically higher), so no arm is failing to fit
the training data. The flat result is not an underfitting symptom.
Second, comparing best-of-five rather than mean-of-five seeds does not
favor the full model. Its best seed ($96.98\%$) is in fact the lowest
best-of-five across all four arms, while the stimulus-only variant's best seed
($97.20\%$) is the highest. If some initializations of the full model
reliably found a materially better configuration, its best-of-five would be
expected to pull ahead of the simpler arms. The observed best-of-five results
therefore favor a genuine performance limit on this benchmark over an
optimization-landscape explanation, in contrast to the bimodal instability
observed on Pubmed. Truncated backpropagation through the
solver, compounded by stacking, may still produce a harder-to-navigate
landscape differently across tasks.

\phantomsection\label{sec:qualdyn}
\paragraph*{Stability dynamics.}

\begin{figure}[h]
\centering
\includegraphics[width=1\textwidth]{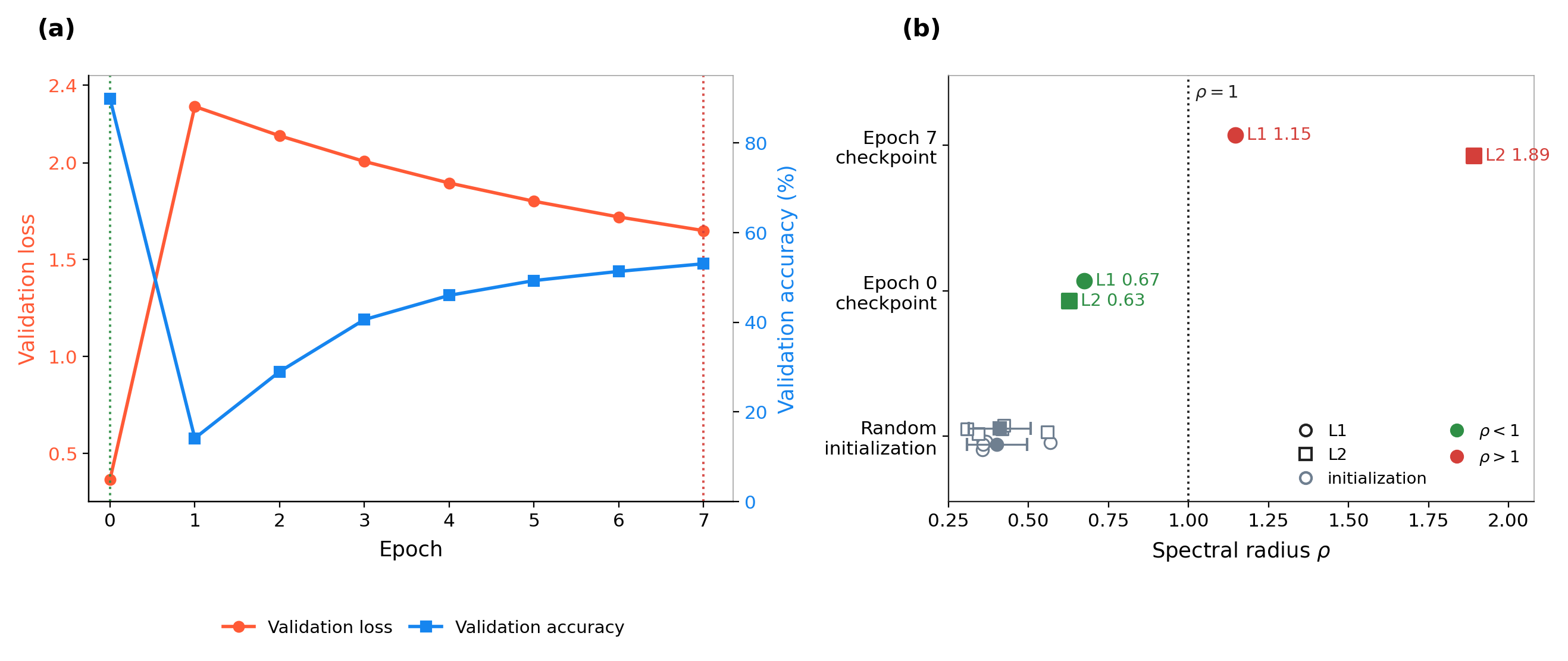}
\caption{Loss of local stability in one adjoint-trained MNIST diagnostic trajectory. (a) Validation
loss and accuracy over training epochs. Dotted vertical lines mark the
checkpoints at which the layer-wise spectral radii were measured.
(b) Spectral radii at random initialization and at the two measured checkpoints.
Circles denote the first equilibrium layer and squares denote the second.
Open gray markers show the five random-initialization seeds for each layer,
and filled gray markers with whiskers show mean $\pm$ standard deviation.
Green checkpoint markers indicate $\rho<1$, red checkpoint markers indicate
$\rho>1$, and the dotted vertical line marks $\rho=1$.}
\label{fig:edgeofchaosreal}
\end{figure}

Figure~\ref{fig:edgeofchaosreal} traces a training instability in
the adjoint-trained MNIST diagnostic described above. In
Figure~\ref{fig:edgeofchaosreal}a, validation accuracy is 89.9\% at the first
saved checkpoint, then collapses to 14.2\% at the following epoch, recovering
only partially to 53.1\% over six subsequent epochs. The paired loss curve remains elevated
after the collapse, indicating that the model did not simply move to a
different high-accuracy solution with changed calibration. The spectral radius
(Algorithm~\ref{alg:rho}) was $\rho_{L1}=0.67$, $\rho_{L2}=0.63$
(below the local $\rho=1$ boundary) immediately before the instability and
$\rho_{L1}=1.15$, $\rho_{L2}=1.89$ (above that boundary) at the final epoch.
Figure~\ref{fig:edgeofchaosreal}b
compares those measured checkpoints with layer-wise random-initialization
values from five seeds for the same configuration. The initialization values lie well below the
$\rho=1$ boundary, whereas the trained checkpoint moves one layer moderately and
the other layer strongly beyond it. The instability is therefore associated with
the training trajectory moving the local Jacobian beyond the measured
$\rho=1$ stability boundary, not with this configuration starting beyond that
boundary at initialization.

\begin{figure}[h]
\centering
\includegraphics[width=1\textwidth]{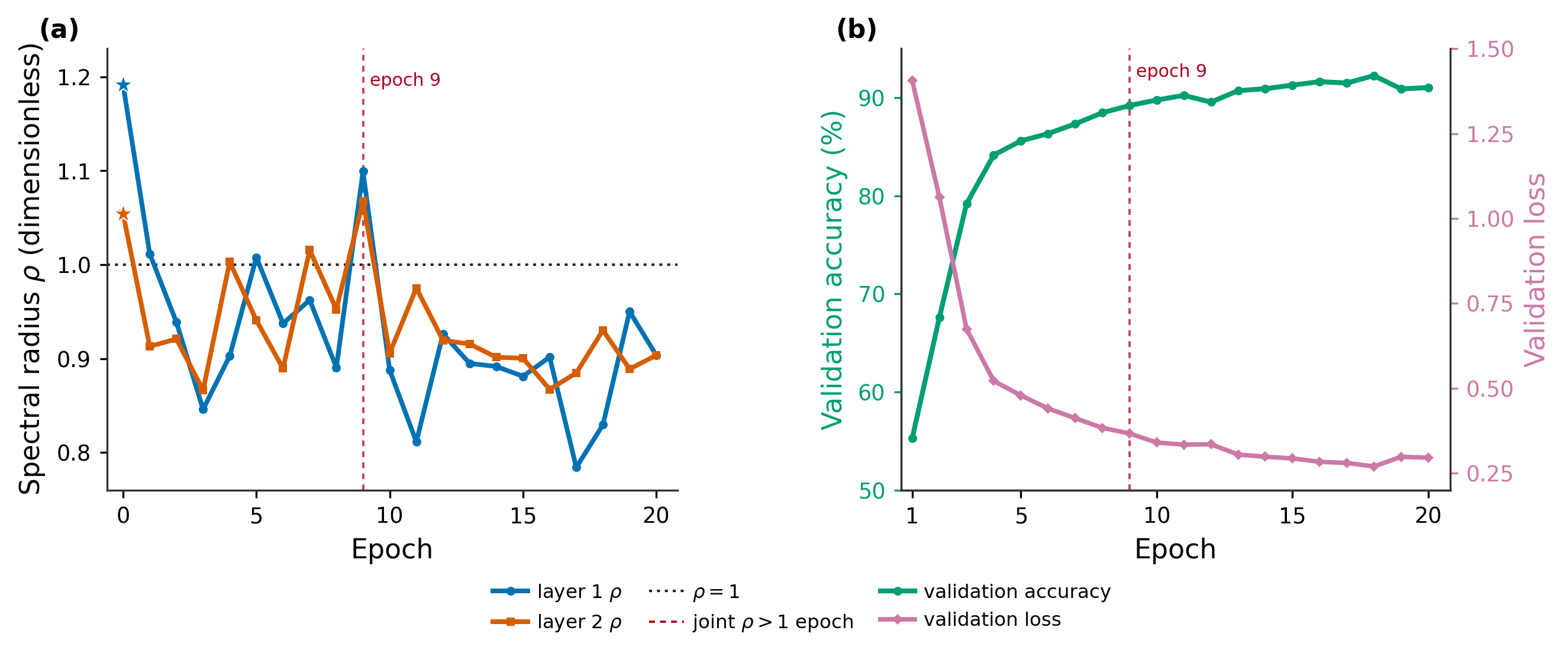}
\caption{Epoch-resolved spectral-radius probe for the configuration-space
grid's largest-\(\rho\) cell at initialization. (a) Layer-wise spectral
radius during training. Epoch 0 denotes random initialization, the horizontal
dotted line marks $\rho=1$, and the red vertical line marks epoch 9, where both
layers briefly exceed $\rho=1$. (b) Validation accuracy and validation loss for
the same trajectory.}
\label{fig:instabilityprobe}
\end{figure}

\paragraph*{Epoch-resolved near-boundary trajectory.}
The configuration-space analysis below
identifies its largest-\(\rho\) cell at random initialization (multi-head
attention local term, node-to-node global-attention term, mean $\rho=1.27$ across five
seeds). One instance of that configuration was trained for twenty epochs with
$\rho$ estimated at every epoch for both layers
(Figure~\ref{fig:instabilityprobe}). This trajectory began with both layers
above $\rho=1$ ($\rho_{L1}=1.19$, $\rho_{L2}=1.05$ at initialization), the opposite
starting condition from the instability traced above. Training nonetheless
converged normally. Validation accuracy rose from 55.3\% (epoch 1) to a peak
of 92.3\% (epoch 18), ending at 91.0\% (epoch 20), while validation loss fell
from 1.41 to 0.30 over the same trajectory. $\rho$ for both layers fell below
1 within the first three epochs and then
oscillated in a narrow band (0.78--1.10) for the remainder of training. It
crossed back above 1 intermittently, in three of twenty epochs for each layer,
including one epoch (epoch 9) where both layers exceeded 1 simultaneously
($\rho_{L1}=1.10$, $\rho_{L2}=1.07$), without any accompanying drop in
validation accuracy.

The epoch-resolved trace shows that crossing $\rho=1$ is not sufficient, by
itself, to trigger the kind of collapse traced above. In
Figure~\ref{fig:instabilityprobe}a, both layers start at or above the
local stability boundary, fall below it within the first few epochs, and make only
brief excursions above 1 later in training. The marked epoch 9 is the strongest
joint excursion, but both layers remain close to the boundary
($\rho_{L1}=1.10$, $\rho_{L2}=1.07$). In
Figure~\ref{fig:instabilityprobe}b, validation accuracy continues rising through
the same epoch, and validation loss continues its downward trend over the
surrounding epochs. Together, the two
trajectories distinguish transient threshold crossings from sustained
large excursions above the measured boundary. The collapsed trajectory was
substantially and persistently above the boundary in the second layer
($\rho_{L2}=1.89$), whereas the probe's
excursions above 1 were small and self-corrected within one to two epochs. The
evidence is consistent with brief, small excursions near $\rho=1$ being
tolerated, while a large, sustained departure from the stability boundary marks the
observed instability. This refines the spectral-radius diagnostic within the
probed configuration.

\phantomsection\label{sec:configspace}
\paragraph*{Configuration-space stability.}

\begin{figure}[h]
\centering
\includegraphics[width=1\textwidth]{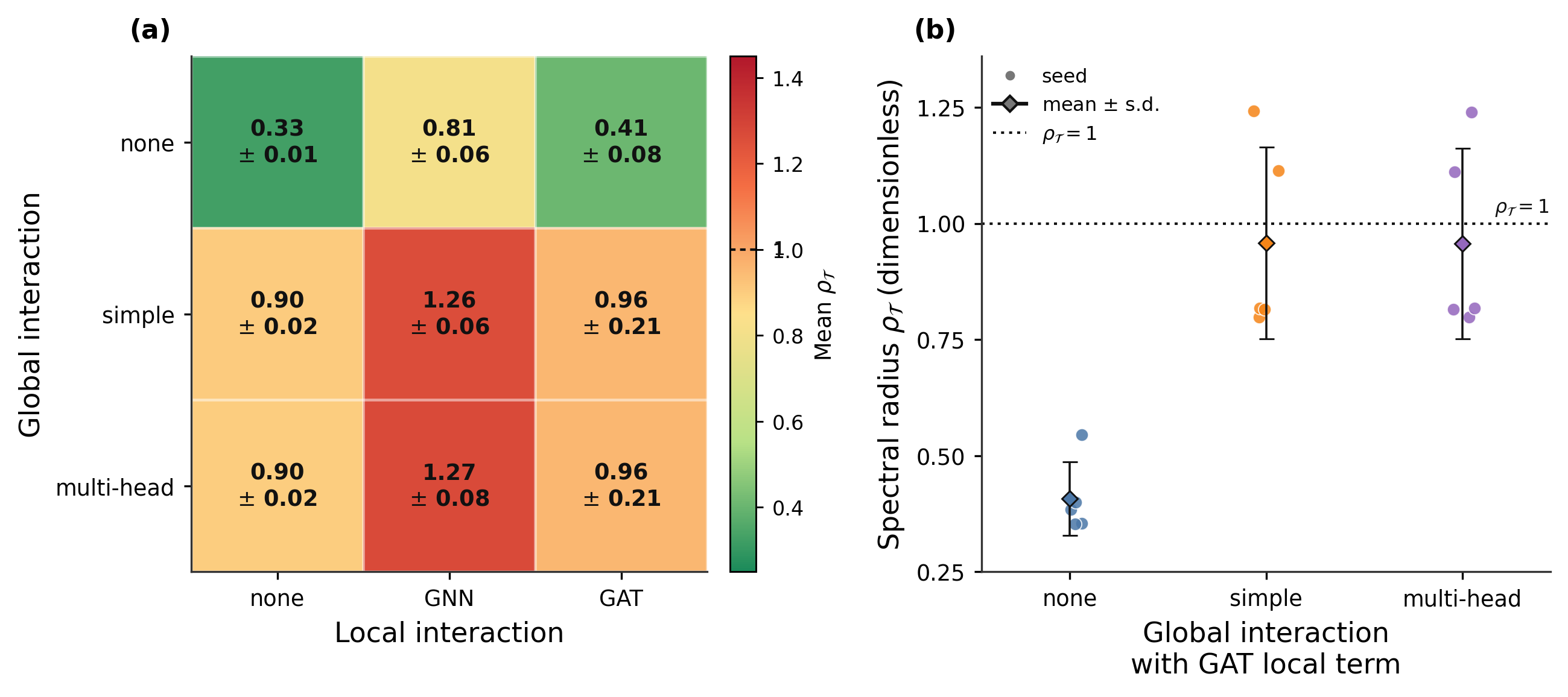}
\caption{Configuration-space map of $\rho_{\mathcal T}$ at random initialization.
(a) Mean $\rho_{\mathcal T}$ for each pairing of global and local interaction
modes. Cells report mean $\pm$ standard deviation over $n=5$ seeds. (b)
Individual seed estimates in the GAT local-interaction column, with diamonds
and whiskers showing mean $\pm$ standard deviation. Dotted lines mark
$\rho_{\mathcal T}=1$.}
\label{fig:configspace}
\end{figure}

Figure~\ref{fig:configspace} maps $\rho_{\mathcal T}$ at random initialization
across the full grid of local- and global-interaction variants, with five seeds
per cell. In Figure~\ref{fig:configspace}a, color encodes the across-seed mean
and each cell reports mean $\pm$ standard deviation. The no-global row remains
below $\rho_{\mathcal T}=1$ throughout the local-interaction choices, including the
configuration underlying the instability traced above
($\rho_{\mathcal T}=0.41\pm0.08$). Adding simple or multi-head global attention
raises the gain substantially. The graph-neural local column crosses the
local stability boundary for both global-attention variants
($\rho_{\mathcal T}=1.26\pm0.06$ and $1.27\pm0.08$), while the GAT local column
moves close to the boundary with larger seed variation
($0.96\pm0.21$). Figure~\ref{fig:configspace}b isolates that GAT local column
and shows the five individual seed estimates for each global-interaction mode.
Changing only the global interaction shifts the values from a safely
below-boundary regime to a near-boundary regime in which some seeds exceed
$\rho_{\mathcal T}=1$. Local stability is therefore not uniform across this
architecture's own configuration space and cannot be assumed by analogy from
one working configuration to a nearby one.

\phantomsection\label{sec:robustnessrho}
\paragraph*{Robustness versus spectral radius.}

\begin{figure}[h]
\centering
\includegraphics[width=1\textwidth]{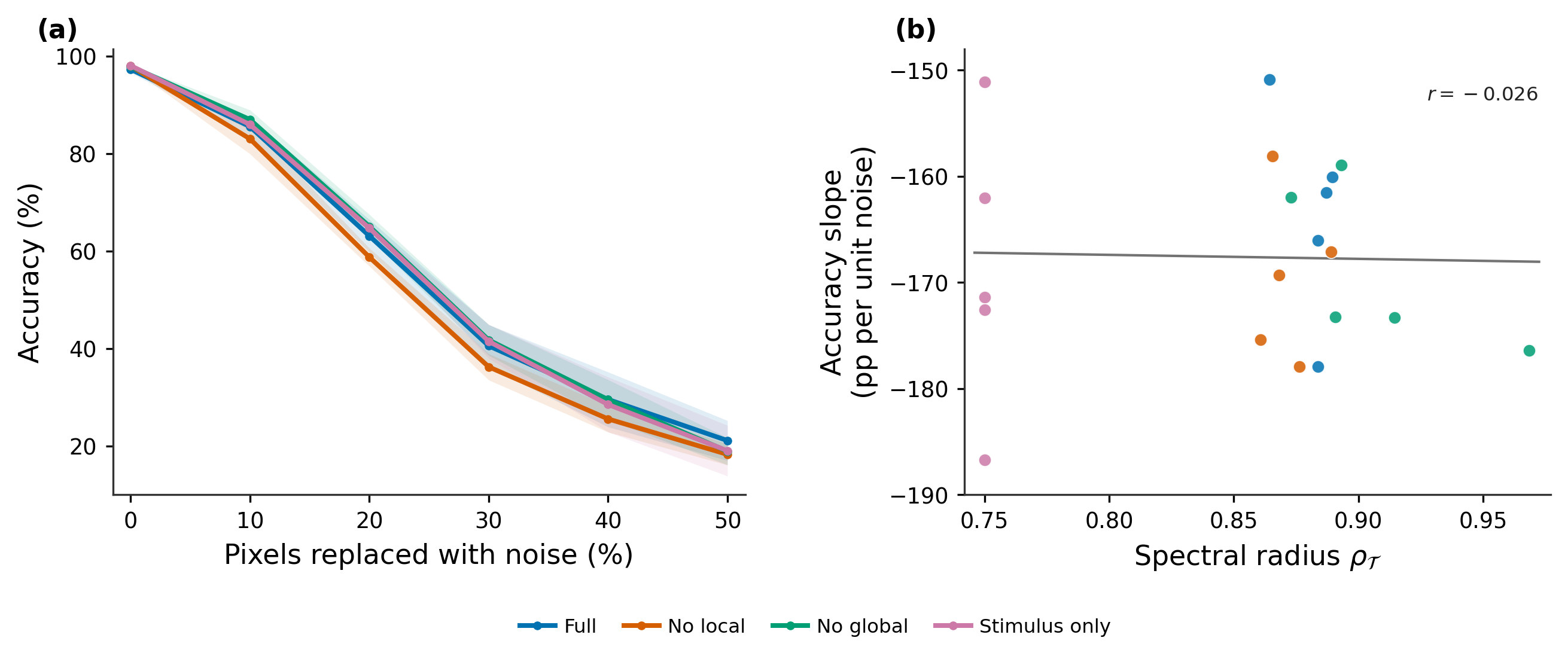}
\caption{Robustness and spectral radius for twenty independently-trained
MNIST ablation-matrix models. (a) Accuracy as increasing fractions of input
pixels are replaced with noise. Lines show means and bands show standard
deviations across five seeds per ablation family. (b) Measured
$\rho_{\mathcal T}$ at each model's converged fixed point versus the fitted
accuracy-degradation slope. Slopes are reported in percentage points per unit
noise fraction, and the gray line shows the least-squares fit.}
\label{fig:robustnessrho}
\end{figure}

Section~\ref{sec:pathsum} and Figure~\ref{fig:edgeofchaosreal} connect
$\rho$ approaching 1 to loss of local linear stability and to one observed
training instability. A broader test evaluates whether a converged equilibrium closer to
$\rho=1$ degrades faster under input noise, with no additional training.
$\rho$ was estimated
(Algorithm~\ref{alg:rho}) at each of twenty independently-trained MNIST
ablation-matrix models' own converged fixed
point, and each model was evaluated with increasing fractions of input
pixels replaced by noise (Figure~\ref{fig:robustnessrho}).

There is no measurable correlation (Pearson $r=-0.026$, $n=20$). In
Figure~\ref{fig:robustnessrho}a, the mean accuracy curves for the four ablation
families degrade smoothly as the fraction of corrupted pixels increases, and
their seed-to-seed bands overlap over most noise levels. No ablation family
separates into a distinctly more robust or more fragile group. In
Figure~\ref{fig:robustnessrho}b, the measured fixed-point spectral radii do not
order the degradation slopes, and individual checkpoints from the four
interaction variants remain intermingled. The estimator itself is calibrated by
the stimulus-only variant, where both interaction terms are disabled. This
variant measures $\rho=0.75$ for all five seeds. It matches the analytically
exact value $(1-\alpha)$ for $\alpha=0.25$ that this configuration's update
provably reduces to, since with both interaction terms off the stimulus does
not depend on the state at all and the true Jacobian is exactly
$(1-\alpha)I$. This null result covers one perturbation type, one dataset, and
a $\rho$ range (0.75--0.97) below the instability threshold observed in
training. The stability-dynamics traces provide the complementary
epoch-resolved view. A system deliberately tracked through the transition
across $\rho=1$ during training, rather than evaluated only at a safely
below-boundary operating point, shows that transient, small crossings of $\rho=1$
are tolerated without collapse, while the unstable trajectory analyzed above
involved a much larger and sustained departure from the local stability
boundary. Both results
support a stability-diagnostic role for $\rho$ in SILVA. Chu et
al.~\cite{chu2024lyapunov} report that \emph{imposing} Lyapunov stability as an
explicit constraint, rather than observing where a network's $\rho$ happens to
fall, yields a robustness benefit for deep equilibrium models.

\section{Discussion}\label{sec:discussion}

The experiments support SILVA as a structured implicit-layer framework in
which input stimulus, local interaction, global context, and self-persistence
are separated as trainable components of the same equilibrium computation. The
model reads out the settled response of an interacting system driven by the
input and relaxed by a damped solver. This separation is the main
architectural contribution. It makes the stimulus entering the system, the
routes through which that
stimulus propagates, the damping that controls relaxation, and the vector
attractor that is ultimately read out available for ablation, visualization,
and stability analysis.

The experiments establish three consequences of this formulation. First, the
same outer equilibrium template transfers across image classification,
molecular property regression, citation-network node classification, and
long-range node classification, while the local and global operators remain
adapted to each domain. Across these settings, the shared object is the
decomposition into stimulus, local interaction, global interaction, and solver,
with each domain specifying the state space and the corresponding interaction
operators. Second, the representation can be inspected through its own
operators. The MNIST results expose learned
neighborhoods, attention weights, global interaction structure, solver
residuals, and equilibrium-state geometry. The spectral-radius and path-sum
derivations connect those measurements to the damped iteration rather than
treating them as post-hoc visualizations. Third, stacked SILVA layers compose
separately solved approximate vector attractor states, giving depth through a sequence of
equilibrium responses rather than through a larger number of ordinary
feedforward transformations.

The empirical pattern is consistent with the intended role of the global term.
On MNIST and citation-network node classification, the additional global
interaction does not improve over a local-only or no-interaction baseline.
These are regimes in which local structure, task simplicity, or limited label
budgets can already determine much of the solution. On CLUSTER, where
long-range aggregation is central \cite{dwivedi2020benchmarking}, the same
decomposition identifies the opposite case. The full local+global mechanism reaches $73.04\pm0.60\%$ across five
independent seeds, exceeding the same-budget gated graph-convolution rerun by
$14.29$ percentage points. A
four-seed no-global ablation reaches $67.55\pm0.84\%$, giving a
$5.49$ percentage-point aggregate difference associated with the global term
in the long-range setting. This contrast identifies the regime in which the extra mechanism
contributes measurable signal.

SILVA is also distinct from the related families that motivate it. Deep
equilibrium models and multiscale DEQs provide the fixed-point view and the
implicit-gradient framework \cite{bai2019deq,bai2020mdeq}. SILVA uses that
view to expose the internal composition of the equilibrium update. Graph neural
networks and attention layers provide learned interaction operators
\cite{kipf2017gcn,velickovic2018gat,vaswani2017attention}. SILVA places those
operators inside a repeated, damped fixed-point field and separates the
external stimulus from local and global interactions. Hopfield and related
attractor models provide useful language for settled states and collective
dynamics \cite{hopfield1982neural}, but SILVA does not require symmetric
weights, a Hopfield energy, or biological interpretation. Graph Perceiver IO
remains a strong reference point for the same citation-network benchmark family
\cite{bae2022graphperceiverio}. SILVA contributes a complementary formulation,
a general implicit interaction field whose operators and dynamics are
explicitly defined and empirically interrogated.

Several points delimit this interpretation. Most reported experiments train the
finite unrolled solver with truncated backpropagation through Picard
iterations. Only the adjoint-trained MNIST diagnostic uses the GMRES adjoint
described in the Supplementary Methods.
Accordingly, the principal reported results characterize weight-tied iterative
networks trained through finite solver trajectories, without an implied
$O(1)$-memory guarantee. Even when an adjoint solve is used, it is computed
numerically by GMRES to a fixed tolerance, so the resulting gradient is an
approximation to an implicit-function-theorem gradient. This is a known source
of instability in DEQs and motivates recent reversible formulations that
compute exact gradients directly \cite{mccallum2025reversible}.

The discrete neighborhoods used by the vision hidden-channel graph and the
node-classification top-$k$ global-attention variant introduce an additional
qualification. The selected graph is recomputed from the current hidden state,
and top-$k$ selection is not everywhere differentiable. Implicit
differentiation for those cases treats the selected graph as frozen at
$z^\ast$. This qualification does not apply to ZINC, where the local graph is
the fixed molecular bond graph. The Pubmed instability and the three-layer
stacking result also have wider uncertainty than the five-seed aggregate
comparisons. Finally, the Banach fixed-point guarantee underlying
Eq.~\eqref{eq:contraction} and the adjoint derivation of
Eqs.~\eqref{eq:IminusT}--\eqref{eq:adjoint} apply to the nonlinear update as
stated. The path-sum decomposition in Eq.~\eqref{eq:pathsum} and the local
descent surrogate in Eqs.~\eqref{eq:energy-surrogate}--
\eqref{eq:energy-descent-picard} are exact for the specified local
linearization, with $\tanh(z)\approx z$ or
$h_{\mathrm{local}}+h_{\mathrm{global}}$ held fixed for the gradient. They
explain the observed solver dynamics through a controlled approximation rather
than proving an exact identity for the full nonlinear system.

Taken together, the results establish SILVA as a structured way to build and
study implicit neural layers. The method turns within-layer interaction into an
explicit modeling object, makes local and global influence separable inside the
solver, and provides diagnostics that connect performance to the dynamics of
the equilibrium computation. The contribution is therefore methodological as
well as empirical. SILVA defines a learnable interaction field whose settled
vector state can be trained, inspected, stacked, and compared across domains.

\section*{Methods summary}

Full architectural, training, and evaluation detail is given in
Section~\ref{sec:method}. The baseline families are graph convolutional
networks (GCNs), graph attention networks (GATs), graph isomorphism networks
(GINs), principal neighbourhood aggregation (PNA), and gated graph convolutional
networks (GatedGCNs). They were trained under the same dataset splits, training
and evaluation protocol, and hidden-width setting as the corresponding SILVA
configuration in each experiment. They are not
parameter-matched unless explicitly stated. Means and standard deviations are computed
over five independent seeds unless stated otherwise.
Spectral radius (Algorithm~\ref{alg:rho}) is estimated by power iteration
on the vector--Jacobian product of the relevant equilibrium layer's update
map, evaluated either at random initialization in the configuration-space
analysis or at the converged fixed point of a trained model in the robustness
analysis, as stated in each case.

Table~\ref{tab:hyperparams} lists the training hyperparameters for SILVA in each
experiment. Baseline hyperparameters follow the corresponding baseline family
under the same training scale, rather than being forced to identical parameter
counts or identical optimizer settings.

\begin{table}[h]
\centering
\caption{Training hyperparameters for SILVA by experiment.
Here $d$ is hidden dimension, $\eta$ is learning rate, $B$ is batch size, and
$k$ is local $k$-nearest-neighbor size for vision or the number of attention
heads for graph experiments. Patience and gradient-clip norm are as defined in
Section~\ref{sec:method}'s Training
paragraph.}
\label{tab:hyperparams}
\setlength{\tabcolsep}{4pt}
\begin{tabular}{@{}p{0.27\textwidth}ccccccc@{}}
\toprule
Experiment & $d$ & $\eta$ & $B$ & Epochs & Pat. & Clip & $k$/heads \\
\midrule
MNIST & 64 & 0.002 & 128 & 30 & 5 & 1.0 & 4 \\
CIFAR-10 & 64 & 0.002 & 128 & 30 & 5 & 1.0 & 4 \\
ZINC & 64 & 0.001 & 128 & 100 & 15 & 1.0 & --- \\
Cora / Citeseer / Pubmed & 64 & 0.01 & $1^{\mathrm{a}}$ & 100 & 20 & 1.0 & 4 \\
CLUSTER & 64 & 0.001 & 32 & 50 & 10 & 1.0 & 4 \\
\bottomrule
\end{tabular}
\par\smallskip
\begin{minipage}{0.92\textwidth}
\footnotesize\raggedright
$^\mathrm{a}$Cora/Citeseer/Pubmed are single-graph transductive tasks. Batch
size is not meaningful and is fixed at 1 (the whole graph) under the evaluation
protocol. Damping is architecture-specific. MNIST uses one equilibrium layer
with $\alpha=0.25$ and 20 Picard iterations. CIFAR-10 and ZINC use two
equilibrium layers with $\alpha_1=0.5$, $\alpha_2=0.2$, and 20 Picard
iterations per layer. Node-classification and graph-benchmark experiments use
two equilibrium layers with $\alpha_1=0.5$, $\alpha_2=0.2$, and 15 Picard
iterations per layer unless an ablation changes the stack, damping, or
local-interaction repetition depth.
\end{minipage}
\end{table}

\section*{Supplementary Methods for scalar and stacked equilibrium derivations}

\subsection*{Three scalar nodes}

This scalar example is the minimal computational version of the
interacting equilibrium layer. Let the input be $x\in\mathbb R^P$, and let
one hidden equilibrium layer contain three nodes. The same receiver-first
convention is used here. The entry $a_{ij}$ is the influence from source
node $j$ to receiver node $i$. The feed-forward stimulus for node
$i\in\{1,2,3\}$ is
\begin{equation}
m_i=\sum_{r=1}^{P}W_{ir}x_r+c_i,
\label{eq:supp-stimulus}
\end{equation}
Here $r$ indexes input features, $W_{ir}$ is the weight from input coordinate
$r$ to hidden node $i$, and $c_i$ is the bias for that node.
Equivalently, in matrix form,
\begin{align}
m &=
\begin{bmatrix}
W_{11} & \cdots & W_{1P}\\
W_{21} & \cdots & W_{2P}\\
W_{31} & \cdots & W_{3P}
\end{bmatrix}x+
\begin{bmatrix}c_1\\c_2\\c_3\end{bmatrix},
\label{eq:supp-stimulus-matrix}\\
W &\in\mathbb R^{3\times P},
\label{eq:supp-W-space}\\
c,m &\in\mathbb R^3.
\label{eq:supp-vector-space}
\end{align}
Here $W$ stacks the three stimulus rows, $c$ stacks the three biases, and
$m=(m_1,m_2,m_3)^\top$ is the stimulus vector injected into the equilibrium
layer.
The intra-layer interactions are
\begin{align}
b_1 &= m_1+a_{11}\phi(b_1)+a_{12}\phi(b_2)+a_{13}\phi(b_3),
\label{eq:supp-three-node}\\
b_2 &= m_2+a_{21}\phi(b_1)+a_{22}\phi(b_2)+a_{23}\phi(b_3),
\label{eq:supp-three-node-second}\\
b_3 &= m_3+a_{31}\phi(b_1)+a_{32}\phi(b_2)+a_{33}\phi(b_3),
\label{eq:supp-three-node-third}
\end{align}
Here $b_i$ is the pre-activation state of node $i$, $\phi$ is the scalar
activation applied elementwise, and each $a_{ij}$ multiplies the nonlinear
signal sent from source node $j$ to receiver node $i$.
The compact vector form is $b=m+A\phi(b)$ with
\begin{align}
A &=
\begin{bmatrix}
a_{11} & a_{12} & a_{13}\\
a_{21} & a_{22} & a_{23}\\
a_{31} & a_{32} & a_{33}
\end{bmatrix},
\label{eq:supp-A}\\
A &\in\mathbb R^{3\times3}.
\label{eq:supp-A-space}
\end{align}
Here $A$ collects all pairwise interaction coefficients, including the diagonal
self-interaction coefficients $a_{11},a_{22},a_{33}$.
The nonlinear SILVA model solves this system by iteration, either as
the undamped map
\begin{equation}
b_{k+1}=m+A\phi(b_k),
\label{eq:supp-undamped-iteration}
\end{equation}
or as the damped map
\begin{equation}
b_{k+1}=(1-\alpha)b_k+\alpha[m+A\phi(b_k)].
\label{eq:supp-damped-iteration}
\end{equation}
Here $b_k$ is the solver iterate after $k$ updates and $\alpha$ is the same
damping coefficient used in Eq.~\eqref{eq:picard}. The bracketed term is the
undamped nonlinear update, and the prefactor $(1-\alpha)b_k$ carries forward
part of the previous state.
In the linear special case $\phi(b)=b$, the exact solution is
\begin{equation}
b^\ast=(I-A)^{-1}m
\label{eq:supp-linear-solution}
\end{equation}
whenever $\rho(A)<1$. Here $I$ is the $3\times3$ identity matrix, $\rho(A)$ is
the spectral radius of $A$, and $b^\ast$ is the converged equilibrium state.

\subsection*{From three nodes to $N$ scalar nodes}

The three-node construction extends directly to an arbitrary number of scalar
state nodes. Let one equilibrium layer contain $N$ nodes and let the encoded
input be $x\in\mathbb R^P$. The stimulus at receiver node $i$ is
\begin{equation}
m_i=\sum_{r=1}^{P}W^{(s)}_{ir}x_r+c_i.
\label{eq:supp-N-stimulus}
\end{equation}
Here $i\in\{1,\ldots,N\}$ indexes state nodes, $r$ indexes input coordinates,
$W^{(s)}\in\mathbb R^{N\times P}$ is the stimulus matrix, and
$c\in\mathbb R^N$ is the stimulus bias.

The full scalar interacting layer is
\begin{equation}
b_i=m_i+\sum_{j=1}^{N}A_{ij}\phi(b_j).
\label{eq:supp-N-node}
\end{equation}
Here $j$ indexes source nodes, $A_{ij}$ is the receiver-first interaction
coefficient from source $j$ to receiver $i$, and $\phi$ is the scalar
nonlinearity. In vector form,
\begin{align}
b &= m+A\phi(b),
\label{eq:supp-N-vector}\\
A &= A_{\mathrm{local}}+A_{\mathrm{global}}.
\label{eq:supp-N-local-global}
\end{align}
The matrix $A_{\mathrm{local}}$ contains the sparse local interactions. Its
entry $(A_{\mathrm{local}})_{ij}$ is zero when $j$ is not in the local
neighborhood of $i$. The matrix $A_{\mathrm{global}}$ contains dense or
low-rank context interactions. For a scalar mean-field broadcast, one example is
\begin{equation}
(A_{\mathrm{global}})_{ij}=\frac{\beta}{N},
\label{eq:supp-N-mean-field}
\end{equation}
where $\beta$ is the scalar global gate. This is the scalar analogue of the
mean-field broadcast in Eq.~\eqref{eq:global-broadcast}.

The damped solver for the $N$-node scalar system is
\begin{equation}
b_{k+1}=(1-\alpha)b_k+\alpha\big[m+(A_{\mathrm{local}}+A_{\mathrm{global}})\phi(b_k)\big].
\label{eq:supp-N-damped}
\end{equation}
Here $b_k\in\mathbb R^N$ is the solver state after $k$ iterations and
$\alpha$ is the damping coefficient. Linearizing around a fixed point replaces
$\phi(b_k)$ by a local linear action, giving the finite-solver transition
\begin{align}
b_{k+1} &= \alpha m+T_\alpha b_k,
\label{eq:supp-N-linear-step}\\
T_\alpha &= (1-\alpha)I+\alpha A,
\label{eq:supp-N-transition}\\
A &= A_{\mathrm{local}}+A_{\mathrm{global}}.
\label{eq:supp-N-A-split}
\end{align}
The identity $I$ has size $N\times N$. Starting from $b_0=0$, repeated
substitution gives
\begin{align}
b_1 &= \alpha m,
\label{eq:supp-N-first-step}\\
b_2 &= \alpha m+\alpha T_\alpha m,
\label{eq:supp-N-second-step}\\
b_3 &= \alpha m+\alpha T_\alpha m+\alpha T_\alpha^2m.
\label{eq:supp-N-third-step}
\end{align}
Therefore, after $K$ iterations,
\begin{equation}
b_K=\alpha\sum_{t=0}^{K-1}T_\alpha^t m.
\label{eq:supp-N-finite-response}
\end{equation}
The term $T_\alpha^t m$ contains every interaction history of length $t$ in
which the stimulus has been repeatedly carried by self-persistence, local
interaction, or global interaction.

If the self-persistence factors are collected separately, the same finite
response can be expressed in powers of the learned interaction matrix $A$,
\begin{align}
b_K &= \sum_{r=0}^{K-1} w_r^{(K,\alpha)} A^r m,
\label{eq:supp-N-pathsum}\\
w_r^{(K,\alpha)}
&= \alpha\sum_{q=r}^{K-1}\binom{q}{r}(1-\alpha)^{q-r}\alpha^r.
\label{eq:supp-N-path-weight}
\end{align}
Here $r$ is the number of local/global interaction applications and
$w_r^{(K,\alpha)}$ is the weight assigned by a finite damped solver to all
length-$r$ interaction histories. Since
$A=A_{\mathrm{local}}+A_{\mathrm{global}}$,
\begin{equation}
A^r=\sum_{w\in\{\mathrm{local},\mathrm{global}\}^r}
A_{w_r}\cdots A_{w_1}.
\label{eq:supp-N-word-expansion}
\end{equation}
The word $w=(w_1,\ldots,w_r)$ records which branch is applied at each
interaction step. This is the scalar multi-node version of the path-sum
interpretation in Section~\ref{sec:pathsum}.

\subsection*{From scalar nodes to vector-feature nodes}

The evaluated graph and molecule layers use vector features at each node. Let
$B\in\mathbb R^{N\times d}$ be the state matrix, with row $B_i\in\mathbb R^d$
holding the $d$ features of node $i$. Let $X\in\mathbb R^{N\times p}$ be the
encoded node input. A vector-feature stimulus can be written
\begin{equation}
S_i=X_iW_{\mathrm{stim}}^\top+c_{\mathrm{stim}}.
\label{eq:supp-vector-stimulus}
\end{equation}
Here $S_i\in\mathbb R^d$, $X_i\in\mathbb R^p$,
$W_{\mathrm{stim}}\in\mathbb R^{d\times p}$, and
$c_{\mathrm{stim}}\in\mathbb R^d$. Let
$Y=\chi_{\mathcal D}(B)$ be the recurrent signal supplied to the interaction
branches. The local and global linearized branches can be written by blocks,
\begin{align}
L(Y)_i &= \sum_{j=1}^{N}A^{L}_{ij}Y_jW_L^\top,
\label{eq:supp-vector-local}\\
G(Y)_i &= \sum_{j=1}^{N}A^{G}_{ij}Y_jW_G^\top.
\label{eq:supp-vector-global}
\end{align}
Here $A^L$ is sparse over the local graph, $A^G$ is dense or low-rank over the
global context, and $W_L,W_G\in\mathbb R^{d\times d}$ mix feature channels.
For ZINC, $Y=B$. For the evaluated vision and graph-node models,
$Y=\tanh(B)$.

Flatten the state by stacking all node-feature vectors into
$z=\mathrm{vec}(B)\in\mathbb R^{Nd}$ and flatten the stimulus as
$s=\mathrm{vec}(S)$. Before differentiating through the map $B\mapsto Y$, the
branch matrices associated with
Eqs.~\eqref{eq:supp-vector-local}--\eqref{eq:supp-vector-global} have
receiver-source blocks
\begin{align}
\bar M_{L,ij} &= A^{L}_{ij}W_L,
\label{eq:supp-vector-local-block}\\
\bar M_{G,ij} &= A^{G}_{ij}W_G.
\label{eq:supp-vector-global-block}
\end{align}
Each block has size $d\times d$. Let
\begin{equation}
D_\chi=\frac{\partial\,\mathrm{vec}(\chi_{\mathcal D}(B))}
{\partial\,\mathrm{vec}(B)}.
\label{eq:supp-vector-Dchi}
\end{equation}
Here $D_\chi$ is the identity for ZINC and the diagonal Jacobian of $\tanh$ for
the evaluated vision and graph-node models. The full local
linearizations with respect to the solver state are
\begin{align}
M_L &= \bar M_LD_\chi,
\label{eq:supp-vector-ML}\\
M_G &= \bar M_GD_\chi.
\label{eq:supp-vector-MG}
\end{align}
The matrices $M_L,M_G\in\mathbb R^{Nd\times Nd}$ are therefore the local
linearizations of the two interaction branches. The linearized
damped vector-feature solver is
\begin{equation}
z_{k+1}\approx \alpha s+\big[(1-\alpha)I+\alpha M_L+\alpha M_G\big]z_k.
\label{eq:supp-vector-damped}
\end{equation}
This is the multi-node, multi-feature version of
Eq.~\eqref{eq:source-self-local-global}. The same finite response follows,
\begin{equation}
z_K=R_{K,\alpha}s,
\label{eq:supp-vector-response}
\end{equation}
with
\begin{equation}
R_{K,\alpha}=\sum_{r=0}^{K-1}w_r^{(K,\alpha)}(M_L+M_G)^r.
\label{eq:supp-vector-green}
\end{equation}
Here $R_{K,\alpha}\in\mathbb R^{Nd\times Nd}$ is the finite response operator
realized implicitly by the solver. The matrix is not stored during the
experiments. Its action is represented by repeated calls to the learned update
map.

\subsection*{Stacked and joint multi-equilibrium form}

The stacked SILVA architecture solves multiple equilibrium layers in sequence.
For layer $\ell\in\{1,\ldots,L\}$, let $N_\ell$ be the number of state nodes,
$d_\ell$ be the hidden width, and
$z_{\ell,k}\in\mathbb R^{N_\ell d_\ell}$ be the flattened state at solver
iteration $k$. The layer has its own stimulus map $s_\ell$, local operator
$M_{\ell,L}$, global operator $M_{\ell,G}$, damping coefficient $\alpha_\ell$,
and solver budget $K_\ell$. Its linearized recurrence is
\begin{align}
z_{\ell,k+1}
&= \alpha_\ell s_\ell(x_\ell)+T_{\ell,\alpha}z_{\ell,k},
\label{eq:supp-stack-layer-step}\\
T_{\ell,\alpha}
&= (1-\alpha_\ell)I_\ell+\alpha_\ell M_\ell,
\label{eq:supp-stack-layer-transition}\\
M_\ell &= M_{\ell,L}+M_{\ell,G}.
\label{eq:supp-stack-layer-operator}
\end{align}
Here $I_\ell$ is the identity on the flattened state space of layer $\ell$.
With zero initialization in each layer, the finite response of layer $\ell$ is
\begin{align}
z_{\ell,K_\ell} &= R_{\ell,K_\ell,\alpha_\ell}s_\ell(x_\ell),
\label{eq:supp-stack-layer-response}\\
R_{\ell,K_\ell,\alpha_\ell}
&= \alpha_\ell\sum_{t=0}^{K_\ell-1}T_{\ell,\alpha}^t.
\label{eq:supp-stack-layer-R}
\end{align}
The connector to the next equilibrium is
\begin{equation}
x_{\ell+1}=\psi_\ell(z_{\ell,K_\ell}).
\label{eq:supp-stack-connector}
\end{equation}
Here $\psi_\ell$ is the inter-layer transformation. In the evaluated stacks it
includes the fixed $\tanh$ connector used between solved equilibria.

For a local sensitivity derivation, linearize the stimulus and connector maps,
\begin{align}
S_\ell &= \frac{\partial s_\ell}{\partial x_\ell},
\label{eq:supp-stack-stimulus-jacobian}\\
C_\ell &= \frac{\partial \psi_\ell}{\partial z_{\ell,K_\ell}}.
\label{eq:supp-stack-connector-jacobian}
\end{align}
Here $S_\ell$ maps perturbations in the layer input to perturbations in the
layer stimulus, while $C_\ell$ maps perturbations in one layer's attractor to
the next layer's input. Perturbations propagate through one solved equilibrium
as
\begin{equation}
\delta z_{\ell,K_\ell}
=R_{\ell,K_\ell,\alpha_\ell}S_\ell\delta x_\ell.
\label{eq:supp-stack-perturb-layer}
\end{equation}
For a two-equilibrium stack,
\begin{equation}
\delta z_{2,K_2}
=R_{2,K_2,\alpha_2}S_2C_1R_{1,K_1,\alpha_1}S_1\delta x_1.
\label{eq:supp-stack-two-response}
\end{equation}
For an $L$-equilibrium stack,
\begin{equation}
\delta z_{L,K_L}
=R_{L,K_L,\alpha_L}S_LC_{L-1}\cdots
R_{2,K_2,\alpha_2}S_2C_1R_{1,K_1,\alpha_1}S_1\delta x_1.
\label{eq:supp-stack-L-response}
\end{equation}
This expression is the multi-node, multi-layer version of the finite response
operator. Each $R_{\ell,K_\ell,\alpha_\ell}$ sums paths inside one equilibrium.
Each connector Jacobian $C_\ell$ transfers the settled vector attractor of one
layer into the stimulus of the next.

The finite unrolled backward pass through a sequential stack applies the same
chain rule layer by layer. Let
$\mathcal T_{\ell,\alpha_\ell}(z_\ell,x_\ell)$ be the damped update map of
layer $\ell$. The reverse recursion inside layer $\ell$ is
\begin{align}
\bar z_{\ell,k}
&=J_{\mathcal T_{\ell,\alpha_\ell}}(z_{\ell,k},x_\ell)^\top
\bar z_{\ell,k+1},
\qquad k=K_\ell-1,\ldots,0,
\label{eq:supp-stack-finite-state}\\
\nabla_{\theta_\ell}\mathcal L
&=\sum_{k=0}^{K_\ell-1}
\left(\partial_{\theta_\ell}\mathcal T_{\ell,\alpha_\ell}(z_{\ell,k},x_\ell)\right)^\top
\bar z_{\ell,k+1}.
\label{eq:supp-stack-finite-param}
\end{align}
Here $\bar z_{\ell,k}$ is the loss adjoint for the $k$th iterate of layer
$\ell$, and $\theta_\ell$ denotes only the parameters of that layer. The
terminal adjoint $\bar z_{\ell,K_\ell}$ contains the readout gradient for the
last layer and the connector contribution $C_\ell^\top\bar x_{\ell+1}$ for
earlier layers.

If each layer is differentiated as an exact implicit equilibrium instead, the
stagewise adjoint solves
\begin{align}
\left(I-J_\ell^\top\right)\lambda_\ell
&=\bar z_\ell^\ast,
\label{eq:supp-stack-implicit-solve}\\
\nabla_{\theta_\ell}\mathcal L
&=\left(\partial_{\theta_\ell}f_{\theta_\ell}^{(\ell)}(z_\ell^\ast,x_\ell)\right)^\top
\lambda_\ell,
\label{eq:supp-stack-implicit-param}\\
\bar x_\ell
&=H_\ell^\top\lambda_\ell.
\label{eq:supp-stack-implicit-input}
\end{align}
Here $J_\ell=\partial f_{\theta_\ell}^{(\ell)}/\partial z_\ell$ and
$H_\ell=\partial f_{\theta_\ell}^{(\ell)}/\partial x_\ell$ are evaluated at
the layer fixed point. The upstream gradient for the previous attractor is
$C_{\ell-1}^\top\bar x_\ell$. These equations describe the implicit analogue
of the sequential stack. The experiments with stacked SILVA layers use the
finite unrolled rule above unless explicitly identified as adjoint-trained.

The same sequential stack can be written as one block fixed-point system for
analysis. Define the joint state
\begin{equation}
\mathbf z=(z_1,z_2,\ldots,z_L).
\label{eq:supp-joint-state}
\end{equation}
The feed-forward-coupled fixed-point equations are
\begin{align}
z_1 &= f_{\theta_1}^{(1)}(z_1,x_1),
\label{eq:supp-joint-first}\\
z_2 &= f_{\theta_2}^{(2)}(z_2,\psi_1(z_1)),
\label{eq:supp-joint-second}\\
z_L &= f_{\theta_L}^{(L)}(z_L,\psi_{L-1}(z_{L-1})).
\label{eq:supp-joint-last}
\end{align}
The omitted middle equations follow the same pattern for
$\ell=3,\ldots,L-1$. Linearizing this joint system gives a block lower
bidiagonal Jacobian,
\begin{equation}
J_{\mathbf F}=
\begin{bmatrix}
J_1 & 0 & 0 & \cdots & 0\\
H_2C_1 & J_2 & 0 & \cdots & 0\\
0 & H_3C_2 & J_3 & \cdots & 0\\
\vdots & \vdots & \vdots & \ddots & \vdots\\
0 & 0 & 0 & H_LC_{L-1} & J_L
\end{bmatrix}.
\label{eq:supp-joint-jacobian}
\end{equation}
Here $J_\ell=\partial f_{\theta_\ell}^{(\ell)}/\partial z_\ell$ is the
within-equilibrium Jacobian, $H_\ell=\partial f_{\theta_\ell}^{(\ell)}/\partial
x_\ell$ is the input-to-update Jacobian, and $C_\ell$ is the connector Jacobian
from Eq.~\eqref{eq:supp-stack-connector-jacobian}. The off-diagonal block
$H_{\ell+1}C_\ell$ is nonzero because layer $\ell+1$ receives its input from the
attractor of layer $\ell$.

The joint implicit response is
\begin{equation}
\delta\mathbf z^\ast=(I-J_{\mathbf F})^{-1}\mathbf S\,\delta x_1,
\label{eq:supp-joint-response}
\end{equation}
where $\mathbf S$ injects the external input perturbation into the first layer's
stimulus block. For the sequential SILVA stack, the lower-bidiagonal structure
of Eq.~\eqref{eq:supp-joint-jacobian} makes this inverse equivalent to the
stagewise response in Eq.~\eqref{eq:supp-stack-L-response}. This is different
from a simultaneous multiscale DEQ, where the off-diagonal coupling blocks are
solved together as one enlarged equilibrium rather than as separately settled
vector attractors.

\phantomsection\label{sec:supp-adjoint-mnist}
\subsection*{Adjoint path for the MNIST equilibrium diagnostic}

The experiments use two backward paths. Finite-solver experiments differentiate
the executed Picard iterations as in
Eqs.~\eqref{eq:finite-backward-state}--\eqref{eq:finite-backward-input}. The
separate adjoint-trained MNIST diagnostic uses a linear adjoint solve at the
terminal equilibrium state of the adjoint-trained model. For one such
layer, let
$m=s_\theta(x)$ be the stimulus and let the pre-activation fixed point be
\begin{equation}
b^\ast=m+L_\theta(\phi(b^\ast))+G_\theta(\phi(b^\ast)),
\label{eq:supp-adjoint-undamped-forward}
\end{equation}
with $\phi(b)=\tanh(b)$. The forward Picard loop used for this diagnostic is
undamped. No factor $\alpha$ appears in the layer's forward recurrence or in
its GMRES backward operator. The layer output is
$B^\ast=\phi(b^\ast)$. If $\bar B=\nabla_{B^\ast}\mathcal L$, then the adjoint
entering the pre-activation equilibrium state is
\begin{equation}
\bar b
=\bar B\odot\big(1-\tanh^2(b^\ast)\big).
\label{eq:supp-adjoint-readout}
\end{equation}
Here $\odot$ is elementwise multiplication, and $\bar b$ has the same shape as
$b^\ast$.

The GMRES solve is applied to the local and global linearized actions,
\begin{align}
\mathcal A(v)
&=v-\mathcal J_L(v)-\mathcal J_G(v),
\label{eq:supp-mnist-adjoint-operator}\\
\mathcal A(u)
&=\bar b.
\label{eq:supp-mnist-adjoint-solve}
\end{align}
Here $v$ and $u$ are flattened state-shaped adjoint vectors. The maps
$\mathcal J_L$ and $\mathcal J_G$ are the local- and global-interaction
linear actions evaluated at the terminal equilibrium state. They are not
materialized as matrices. For the static global matrix in the row-oriented
convention, the global action has coordinates
\begin{equation}
\mathcal J_G(v)_i=\sum_{j=1}^{N}A_{ji}\,v_j\,\phi'(b^\ast_j).
\label{eq:supp-mnist-global-action}
\end{equation}
Here $A$ is the learned global interaction matrix, $i$ indexes the output
coordinate of the linear action, and $j$ indexes the source coordinate in the
summation. For the local attention branch, $\mathcal J_L(v)$ is obtained by
distributing the incoming per-node adjoint evenly over the $H$ attention heads,
as in Eq.~\eqref{eq:head-average-adjoint}, and then applying the local
attention layer's backward map.

The solution $u$ is the adjoint used to accumulate the diagnostic gradients.
For a static global matrix, the per-example matrix sensitivity is
\begin{equation}
\widehat{\nabla_A\mathcal L}_{ij}
=\phi(b^\ast_i)u_j.
\label{eq:supp-mnist-A-gradient}
\end{equation}
For a stimulus matrix $W$ and bias $c$, the direct gradients are
\begin{align}
\widehat{\nabla_W\mathcal L}
&=X^\top(u+d_m^{\mathrm{attn}}),
\label{eq:supp-mnist-W-gradient}\\
\widehat{\nabla_c\mathcal L}
&=\sum_{n}(u_n+d_{m,n}^{\mathrm{attn}}).
\label{eq:supp-mnist-c-gradient}
\end{align}
Here $X$ is the minibatch input to the layer, $n$ indexes minibatch elements,
and $d_m^{\mathrm{attn}}$ is the additional stimulus adjoint obtained when the
global interaction matrix is generated by attention from $m$. The hat on these
gradients distinguishes this diagnostic adjoint path from the formal
fixed-point gradient in Eqs.~\eqref{eq:adjoint}--\eqref{eq:implicit-input-gradient}.
The conditioning of such forward and backward linearized solves is controlled
by the equilibrium Jacobian, motivating both the
spectral-radius diagnostics used here and Jacobian-regularized DEQ training in
related work \cite{bai2021jacobian}.

\subsection*{Readout after the equilibrium state}

For binary classification in the three-node scalar guide, the readout can be
written
\begin{align}
\hat y &= \sigma_{\mathrm{sigmoid}}(v^\top\phi(b^\ast)+d),
\label{eq:supp-readout}\\
v &\in\mathbb R^3,
\label{eq:supp-readout-vector}\\
d &\in\mathbb R,
\label{eq:supp-readout-bias}
\end{align}
Here $\hat y$ is the predicted probability for the positive class, $v$ is the
readout vector, $d$ is the scalar readout bias, and
$\sigma_{\mathrm{sigmoid}}$ is the logistic sigmoid. The model is trained with
binary cross-entropy. The multi-class experiments in the
main text use the same principle with a linear map followed by softmax,
applied to the approximate fixed point returned by the finite solver.

\backmatter

\section*{Acknowledgements}

The present work introduces the SILVA Networks methodology and its structured
implicit interaction field. The research program from which this
study emerged was initiated within the Swedish National Infrastructure for
Computing (SNIC) Small Compute project \emph{Artificial Intelligence for
Physics and Engineering, Modeling and Simulation}, Project No. SNIC
2022/22-843. The project was conducted at Link{\"o}ping University under the author's
principal investigatorship. That project established a broader research
direction connecting artificial intelligence, graph neural networks, deep
reinforcement learning, neuromorphic computing, and the modeling and simulation
of physics and engineering systems.

This work was supported by the Brazilian National Council for Scientific and
Technological Development (CNPq) under grant No. 445344/2024-5. The author also
acknowledges financial support provided through the Program Talentos Brasil, as
Project Investigator.

\section*{Declarations}

\begin{itemize}
\item \textbf{Funding.} This work was supported by the Brazilian National Council for
Scientific and Technological Development (CNPq) under grant No. 445344/2024-5.
The author acknowledges financial support provided through the Program Talentos
Brasil, as Project Investigator.
\item \textbf{Competing interests.} The author declares no competing interests.
\item \textbf{Ethics approval.} Not applicable.
\item \textbf{Data and code availability.} MNIST \cite{lecun1998gradient},
CIFAR-10 \cite{krizhevsky2009cifar}, ZINC \cite{irwin2012zinc}, Cora,
Citeseer, Pubmed \cite{sen2008collective,yang2016revisiting}, and CLUSTER
\cite{dwivedi2020benchmarking} are public benchmarks. Exact
split and preprocessing details are given in Section~\ref{sec:method} and the
Methods summary. The computational implementation used Python with standard
scientific-computing and deep-learning software. SILVA Networks research
materials, processed experiment outputs, and additional supporting materials
will be available from the corresponding author upon reasonable request.
\item \textbf{Author contributions.} J.L.L.J.S. conceived the research direction,
developed the SILVA Networks model and benchmark suite, designed and
executed the experiments, interpreted the results, and prepared the manuscript.
\end{itemize}

\bibliographystyle{unsrt}
\bibliography{references}

\end{document}